\documentclass{article}

\usepackage[final]{corl_2026}         %

\makeatletter
\renewcommand{\@conferencelocation}{Austin, Texas, USA}
\makeatother
\hypersetup{
  pdftitle={Noise-Invariant Spatial Encoding for 3D Manipulation Policies under Depth Corruption},
  pdfauthor={Zhiyuan Yang, Jiaju He, Ruiteng Zhao, Anqi Cheng, Haiyue Zhu, Kezhi Mao}
}

\usepackage{graphicx}      %
\usepackage{amsmath}       %
\usepackage{amssymb}       %
\usepackage{booktabs}      %
\usepackage{multirow}      %
\usepackage{cleveref}      %
\usepackage{subcaption}    %
\usepackage{placeins}      %
\usepackage{wrapfig}       %
\usepackage{needspace}     %

\newcommand{\method}{Noise-Invariant Spatial Encoding}

\title{Noise-Invariant Spatial Encoding for 3D Manipulation Policies under Depth Corruption}

\author{
  \textbf{Zhiyuan Yang}$^{1,2}$ \qquad \textbf{Jiaju He}$^{1,2}$ \qquad \textbf{Ruiteng Zhao}$^{1,3}$ \\
  \textbf{Anqi Cheng}$^{2}$ \qquad \textbf{Haiyue Zhu}$^{1,\dagger}$ \qquad \textbf{Kezhi Mao}$^{2,\dagger}$ \\[0.5ex]
  {\normalfont{}$^{1}$Singapore Institute of Manufacturing Technology (SIMTech),} \\
  {\normalfont{}Agency for Science, Technology and Research (A*STAR), Singapore} \\
  {\normalfont{}$^{2}$Nanyang Technological University (NTU), Singapore} \\
  {\normalfont{}$^{3}$National University of Singapore (NUS), Singapore}
}

\begin{document}
\maketitle
\vspace{-29pt}
{\centering\normalfont{}Project page: \url{https://nise-robotics.github.io/}\par}
\vspace{11pt}
\begingroup
\renewcommand{\thefootnote}{\fnsymbol{footnote}}
\footnotetext[2]{Corresponding authors: \href{mailto:zhu_haiyue@a-star.edu.sg}{\nolinkurl{zhu_haiyue@a-star.edu.sg}}; \href{mailto:ekzmao@ntu.edu.sg}{\nolinkurl{ekzmao@ntu.edu.sg}}.}
\endgroup

\begin{abstract}
  3D manipulation policies derive actions from explicit scene geometry rather than color images alone, enabling them to reason over metric distances, object relations, and robot-scene reachability.
  This geometric interface is also a vulnerability: commodity RGB-D sensors return missing measurements, boundary artifacts, and plausible but wrong surfaces around transparent, reflective, thin, or textureless objects.
  We study this failure under realistic depth corruption and observe a sharp depth-noise cliff across modern 3D policies, even when they are trained with matched noisy depth.
  This residual gap motivates a spatial-encoding approach: corrupted depth becomes entangled with the token coordinates and geometry features consumed by the action expert.
  We propose Noise-Invariant Spatial Encoding (\textsc{NISE}), a 3D policy architecture that learns trustworthy geometry features even facing imperfect depth.
  \textsc{NISE} uses Geometric Cross-View Self-Supervision to shape shared policy features with calibrated multi-view depth, and Gaussian Marginalized 3D Attention to carry position uncertainty into deployed attention rather than treating every lifted token as exact.
  Across simulated benchmarks with sensor-realistic corruption and real-robot tasks involving transparent and reflective materials, \textsc{NISE} improves depth-noise robustness while preserving clean-depth performance, supporting 3D policies designed for commodity RGB-D deployment rather than clean-depth benchmarks alone.
\end{abstract}

\section{Introduction}
\label{sec:intro}
\begin{figure}[!h]
  \centering
  \includegraphics[width=1.\linewidth]{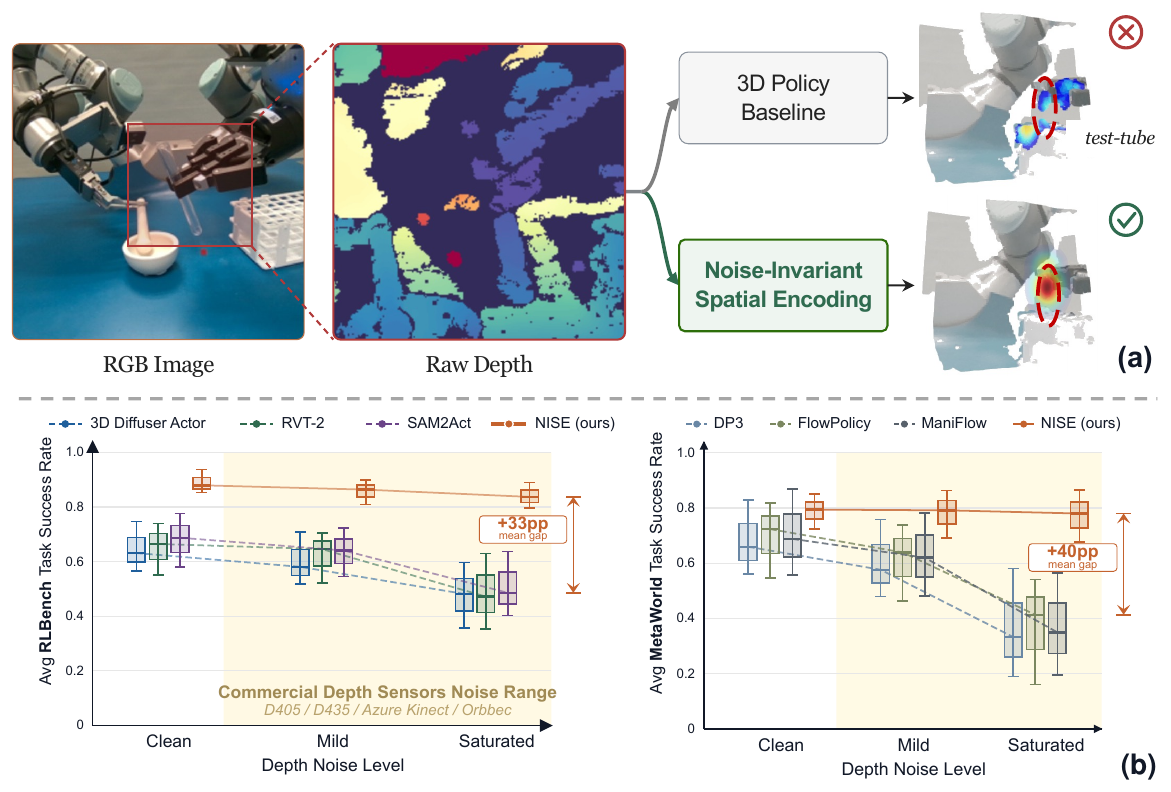}
  \caption{\textbf{Realistic depth corruption breaks what 3D policies can trust.}
  \textbf{(a)} In a transparent test-tube manipulation scene, raw depth contains missing and geometrically wrong measurements; a standard 3D policy consumes this corrupted geometry as exact and fails, while \textsc{NISE} succeeds.
  \textbf{(b)} Depth-noise robustness on RLBench PerAct-18~\citep{rlbench,peract} and MetaWorld~\citep{metaworld}.
  \textsc{NISE} maintains higher success rates as depth corruption increases, with a smaller clean-to-saturated decline than the compared baselines.}
  \label{fig:teaser}
\end{figure}
3D manipulation policies derive actions from explicit scene geometry rather than color images alone, whether by processing point clouds~\citep{dp3,lift3d}, encoding voxel grids~\citep{peract}, reasoning over virtual views~\citep{rvt2,sam2act}, or pairing visual tokens with 3D coordinates~\citep{tddda}.
The appeal comes from the metric nature of manipulation, where contact, clearance, and reachability depend on detecting distances between the robot and the scene.
Depth gives policies direct access to these relationships, turning image observations into spatial representations that can guide action.
This reliance creates a fragile premise: the depth measurements must be reliable enough to support trustworthy geometry.
Commodity RGB-D sensors routinely violate this premise through missing depth measurements, boundary artifacts, and plausible but incorrect surfaces, especially around transparent, reflective, thin, or textureless objects.
This motivates our central question: how severely does realistic depth corruption degrade current 3D policies, and how can 3D policies learn spatial representations that remain reliable under such noise?

We study this question by introducing realistic depth corruption into modern 3D policies.
The failure is sharper than ordinary robustness degradation.
Clean-trained policies deteriorate rapidly as depth corruption increases.
Even after training on matched noisy depth, substantial degradation remains (Fig.~\ref{fig:teaser}).
This suggests that the problem is not simply a missing training augmentation.
These policies construct spatial representations from corrupted depth measurements, so residual depth errors become entangled with the features passed to the action generator.
The key step is therefore to prevent depth corruption from being inherited by the spatial encoding.

Prior work on depth-noise robustness has mostly intervened upstream of the policy, repairing the depth signal before it is consumed by the policy~\citep{liu2025cdm,wei2024d3roma}.
These directions are valuable, but they remain observation-side repairs.
Even with improved geometric representations, residual errors in depth-derived coordinates can continue to perturb positional attention and the action features it produces.
The missing piece is therefore the spatial encoding itself: after depth is sensed, or repaired, the policy still needs a learned ability to form trustworthy geometry features for action prediction.

To address this gap, we propose Noise-Invariant Spatial Encoding (\textsc{NISE}) for 3D manipulation policies.
Rather than treating depth corruption as an input-repair problem, \textsc{NISE} shifts the robustness target to the spatial representation consumed by the action policy.
It introduces two complementary mechanisms.
Geometric Cross-View Self-Supervision (GCVSS) trains the policy encoder with a cross-view geometry head that uses multi-view depth to encourage view-consistent spatial features.
Gaussian Marginalized 3D Attention (GM3A) modifies 3D positional attention with a learned dampening surface, so unreliable coordinate bands have less influence on action-relevant spatial features.

Our contributions are three-fold:
\begin{itemize}
  \item We identify and analyze a depth-noise cliff in 3D manipulation policies, showing that realistic depth corruption can cause sharp performance collapse across visual backbones and action-generation mechanisms, while matched noisy-depth training only partially recovers the loss.
  \item We propose \textsc{NISE}, a 3D policy architecture that learns noise-invariant spatial encodings through Geometric Cross-View Self-Supervision and Gaussian Marginalized 3D Attention, rather than repairing the depth observation.
  \item We evaluate \textsc{NISE} on simulated benchmarks with realistic depth corruption and real-world manipulation tasks involving challenging depth-sensing materials, demonstrating that 3D policies can be designed for sensor-realistic deployment.
\end{itemize}

\section{Related Work}
\label{sec:related-work}

\noindent\textbf{3D Manipulation Policies.}
Explicit 3D representations support both point-cloud recognition and robotic manipulation~\citep{peract,dp3,rvt2,t3dnet}.
Point-cloud policies use explicit 3D geometry to condition action generation~\citep{dp3,lift3d,dream_to_action}; virtual-view policies such as RVT-2~\citep{rvt2} and SAM2Act~\citep{sam2act} render RGB-D observations into canonical views; and depth-lifted-token policies such as 3D Diffuser Actor (3DDA)~\citep{tddda} and 3D FlowMatch Actor~\citep{tdfa} pair visual features with 3D coordinates before action prediction.
These methods differ in action heads and backbones, but they share a useful interface: depth is converted into spatial tokens that the policy can attend to, sample, or aggregate.
Auxiliary reconstruction and distillation objectives provide additional supervision for learning geometric and physical representations~\citep{gnfactor,threedmvp,fdvla}.

A parallel trend strengthens this interface with larger geometric backbones.
R3DP~\citep{r3dp}, VGGT-DP~\citep{vggtdp}, and GP3~\citep{gp3} draw 3D features from frozen geometry transformers, while Spatial Forcing~\citep{spatialforcing} distills VGGT features into a separate action stream.
PromptDA-style depth fusion~\citep{promptda} and 3D rotary position embeddings~\citep{tddda,tdfa} further make metric scene structure available to downstream attention.
Robo3R~\citep{robo3r} reconstructs metric scene geometry from RGB images and robot states for downstream manipulation.
The gap is that these policies usually treat the lifted coordinates as reliable once they enter the model.
Our work asks how the spatial encoder should behave when the depth that constructs those coordinates is uncertain.

\noindent\textbf{Depth Noise and Robustness in Robotic Manipulation.}
Depth robustness is often approached before the policy reasons over the scene.
Classical filters and learned depth completion or estimation methods aim to recover a cleaner depth image~\citep{marigold_dc,depth_anything_3,hfd_teacher}.
Manipulation-specific systems such as CDM~\citep{liu2025cdm} and D3RoMa~\citep{wei2024d3roma} follow the same observation-side logic: improve the depth signal before it is consumed by the policy.
This direction is valuable, but its objective is still depth repair rather than learning how the policy should encode imperfect geometry.

Recent work also addresses noisy geometry through the policy's internal representations.
SpatialActor~\citep{spatialactor} separates semantic and geometric representations and fuses measured-depth features with pretrained depth-expert features.
EquiForm~\citep{equiform} combines geometric denoising with contrastive equivariant representation learning, while PEAfowl~\citep{peafowl} uses depth-distribution lifting, cross-view aggregation, and training-only depth distillation.
These approaches advance robustness through geometric priors, denoising, and depth-aware representation learning.
Our focus is how residual depth errors affect both learned geometric features and the positional interactions between them.
Improving the features alone does not ensure that depth-derived coordinates are reliable for 3D attention.
\textsc{NISE} addresses these coupled effects by learning geometric consistency directly from noisy camera views and adapting the frequency components of 3D rotary attention to coordinate uncertainty.

\section{\texorpdfstring{\textsc{NISE}: Noise-Invariant Spatial Encoding}{NISE: Noise-Invariant Spatial Encoding}}
\label{sec:method}
\begin{figure}[t]
  \centering
  \includegraphics[width=\linewidth]{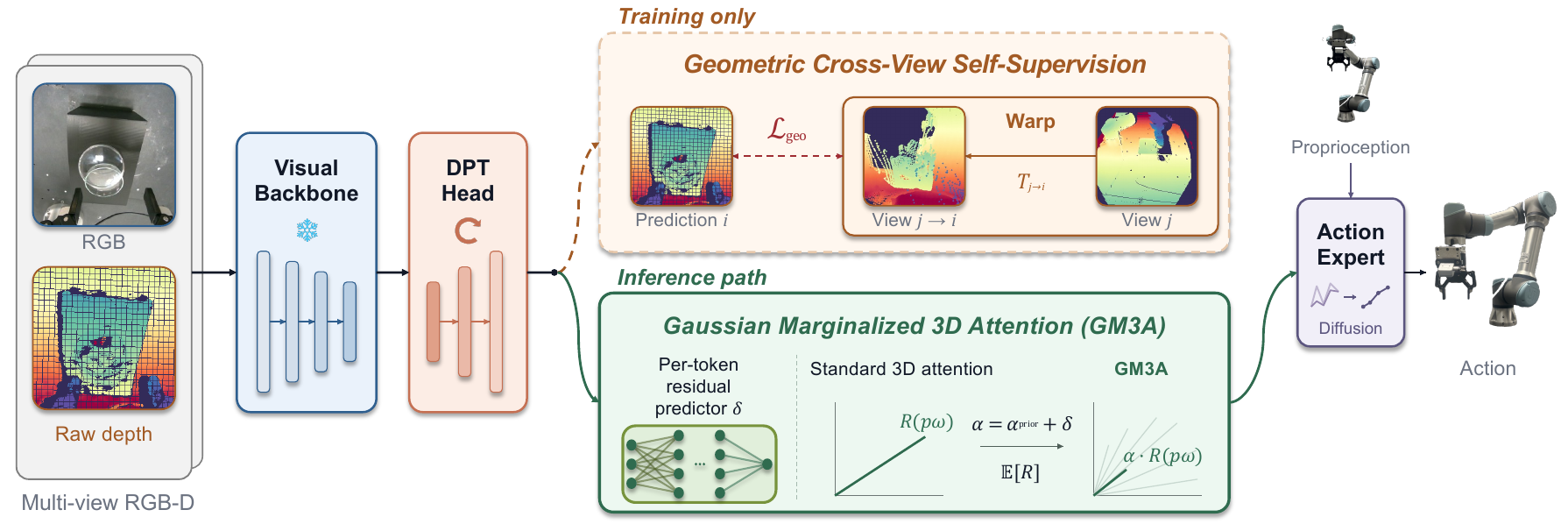}
  \caption{\textbf{\textsc{NISE} framework.}
  Per-camera RGB-D is encoded by a frozen visual backbone ViT and a trainable DPT-style head~\citep{promptda} into a shared spatial feature map.
  \textsc{NISE} acts on this shared map: during training, Geometric Cross-View Self-Supervision attaches a cross-view geometry head to encourage view-consistent geometry in the same features consumed by the policy; at deployment, this head is removed.
  Gaussian Marginalized 3D Attention remains in the deployed policy and carries per-token coordinate uncertainty into 3D attention, so lifted tokens influence action features according to their spatial reliability.}
  \label{fig:method}
\end{figure}

\textsc{NISE} is an architecture for learning \emph{\method{}}: spatial features that remain useful when the depth measurements used to construct them are imperfect.
The design follows two questions.
First, if clean depth is unavailable on real robots, what supervision can teach the policy representation to prefer consistent geometry?
Second, if 3D attention receives coordinates from corrupted depth, how should it reason over positions that are uncertain rather than exact?
Geometric Cross-View Self-Supervision answers the first question by using calibrated multi-view depth as a training signal for the shared policy features.
Gaussian Marginalized 3D Attention answers the second by carrying position uncertainty into the 3D attention.
Together, they move robustness inside the spatial encoding when facing imperfect depth.
\Cref{fig:method} summarizes the architecture.

\subsection{From Depth Noise to Action Error}
\label{sec:method-problem}
We first formulate how depth noise enters a 3D policy and propagates into action error.
For the 3D policies considered here, depth is used together with color to construct spatial observation $\mathcal{O}_{\text{rgbd}}$.
\begin{equation*}
  \mathcal{O}_{\text{rgbd}} = {\{I^{v},\, D^{v},\, K^{v},\, T^{v}\}}_{v=1}^{V},
  \qquad
  a = \pi_\theta(\mathcal{O}_{\text{rgbd}},\, \mathbf{s}_{\text{prop}}).
\end{equation*}
Here $V$ is the number of views, $I^{v}$ and $D^{v}$ are the per-view RGB image and depth map, $K^{v}, T^{v}$ are the per-view intrinsics and extrinsics, and $\mathbf{s}_{\text{prop}}$ denotes proprioception.
The action $a$ may be represented as a keypose, action chunk, or trajectory depending on the action expert; our analysis focuses on how corrupted depth enters the spatial representation before that output stage.
At deployment, the observed depth is corrupted relative to the underlying scene geometry,
\begin{equation*}
  \tilde D^{v} = D^{v} + \epsilon^{v},
\end{equation*}
where $\epsilon^{v}$ denotes depth corruption.
Commodity RGB-D sensors return missing depth, boundary artifacts, and plausible but wrong measurements, especially on transparent, reflective, or textureless surfaces.
For a 3D policy, the damaging step is that this corruption is inherited when scene tokens are assigned 3D coordinates.
Each scene token receives a coordinate by back-projecting its source pixel $(u,v)$ through the observed depth,
\begin{equation*}
  \mathbf{p}(u,v) \;=\; \tilde D(u,v) \cdot K^{-1}{[\,u,\,v,\,1\,]}^{\top},
\end{equation*}
so any depth perturbation becomes a position perturbation, aligned mainly with the camera ray and coupled laterally through the ray Jacobian.
Downstream 3D attention then treats this coordinate as exact.
The error is no longer only a bad depth pixel; it becomes a wrong relative-position phase in 3D rotary attention~\citep{tddda}, a wrong cross-camera spatial relation, and eventually a wrong action feature.

This failure channel explains why broader noise exposure can help yet remain incomplete.
A noise-trained policy sees corrupted observations, but its attention layers still inherit coordinates treated as certain.
\textsc{NISE} therefore asks the encoder to produce a feature
\begin{equation*}
  z_i \;=\; f_\theta\big(I^{v},\, \tilde D^{v},\, \mathbf{p}_i\big)
\end{equation*}
whose value remains useful for action prediction when $\mathbf{p}_i$ is uncertain.
The rest of the section describes how this property is induced during training and preserved at inference.

\subsection{Geometric Cross-View Self-Supervision}
\label{sec:method-gcvss}

The first design problem is where the spatial representation should get its geometric supervision.
Clean simulator depth would be a convenient target, but it would also make the method depend on supervision that is unavailable on real robots.
GCVSS instead uses a signal the robot can collect: calibrated views of the same scene from its own RGB-D cameras.
The cross-view geometry head is attached only during training, so the shared policy features are shaped by geometry that agrees across views without requiring clean depth at deployment.

For target camera $i$, let $j$ denote a paired source camera.
We use calibrated cross-view geometry inspired by stereo depth estimation~\citep{godard2017monodepth,kendall2017gcnet,chang2018psmnet} to forward-warp the noisy source depth $d_j=\tilde D_j$ into camera $i$:
\begin{equation*}
  \mathbf{x}_i
  \sim
  K_i T_{i \leftarrow j}
  \left(
    d_j(\mathbf{x}_j) K_j^{-1} \bar{\mathbf{x}}_j
  \right),
  \qquad
  d_{j \rightarrow i}(\mathbf{x}_i)
  =
  {\left[
    T_{i \leftarrow j}
    \left(
      d_j(\mathbf{x}_j) K_j^{-1} \bar{\mathbf{x}}_j
    \right)
  \right]}_{z} .
\end{equation*}
Here $\bar{\mathbf{x}}_j={[u_j,v_j,1]}^\top$, and $T_{i\leftarrow j}$ maps camera-$j$ coordinates to camera $i$.
The symbol $\sim$ denotes homogeneous projection.
The construction is Noise2Noise-style in spirit~\citep{noise2noise}: one noisy observation provides supervision for another.
The important difference is that the target is geometric and visibility-filtered rather than a strict pixel-wise independent-noise target.
Two views of the same surface may disagree in view-specific sensor artifacts, but after calibration and visibility filtering they agree more consistently on the underlying scene geometry.

Let $\mathbf{F}_i$ be the shared feature map for target camera $i$ and $g_\phi$ be the cross-view geometry head.
We train it with a masked $L_1$ loss over pixels where the source depth is validly splatted into the target view:
\begin{equation*}
  \widehat D_i = g_\phi(\mathbf{F}_i),
  \qquad
  \mathcal{L}_{\mathrm{geo}}
  =
  \frac{1}{|M_{j\rightarrow i}|}
  \sum_{\mathbf{x}\in M_{j\rightarrow i}}
  \left|
    \widehat D_i(\mathbf{x}) - d_{j\rightarrow i}(\mathbf{x})
  \right| .
\end{equation*}
The mask $M_{j\rightarrow i}$ keeps only pixels that land in the target view and pass the visibility test.
The predicted depth map is not used as an external preprocessor.
At inference, the cross-view geometry head is removed; its role is to make the deployed policy features encode geometry that is stable across views rather than artifacts that are idiosyncratic to a single noisy depth image.

\subsection{Gaussian Marginalized 3D Attention}
\label{sec:method-gm3a}

Cross-view supervision shapes the feature map, but it does not change the second place where depth noise enters the policy: the 3D positional attention itself.
Depth-lifted 3D attention usually treats each token coordinate as a point estimate.
This is brittle under corrupted depth because the coordinate controls the phase of 3D rotary position embedding; a small spatial error can become a large high-frequency phase error before the action expert ever sees the features.
The method therefore needs attention to know when geometry should be used precisely and when it should be softened.

GM3A uses a local Gaussian approximation to construct an analytic attenuation prior for uncertain token coordinates.
Averaging sinusoidal positional features under this approximation yields a closed-form, frequency-dependent attenuation~\citep{mipnerf}.
For axis $a$, we approximate $\tilde p_a\sim\mathcal{N}(\mu_a,\sigma_a^2)$; then, for frequency band $k$,
\begin{equation*}
  \mathbb{E}\!\left[e^{\mathrm{i}\omega_k\tilde p_a}\right]
  =
  e^{\mathrm{i}\omega_k\mu_a}
  \underbrace{\exp\!\big(-\sigma_a^2\omega_k^2/2\big)}_{\alpha^{\text{prior}}_{k,a}},
\end{equation*}
which keeps the deterministic phase at the mean coordinate while dampening frequency bands that are fragile to position uncertainty.
Under this prior, larger coordinate uncertainty $\sigma_a$ gives smaller $\alpha^{\text{prior}}_{k,a}$, with stronger attenuation at higher frequencies $\omega_k$.
This attenuation acts on positional components while preserving the tokens' visual features.
For small coordinate uncertainty, the prior approaches one, recovering ordinary 3D-RoPE~\citep{tddda}.

The remaining design choice is how to adapt attenuation to action prediction.
The analytic prior captures the expected effect of sensor uncertainty on positional frequencies, but does not learn which frequency components are most useful for the policy.
GM3A therefore learns a per-token residual around this prior through the policy objective, while regularization keeps the coefficients anchored to sensor-scale uncertainty:
\begin{equation*}
  \alpha_{k,a} = \alpha^{\text{prior}}_{k,a} + \delta_{k,a},
  \qquad
  \mathcal{L}_{\text{reg}}
  =
  \mathbb{E}_{\text{token},k,a}
  \!\left[{(\alpha_{k,a}-\alpha^{\text{prior}}_{k,a})}^{2}\right].
\end{equation*}
The learned residual allows the coefficients to depart from the Gaussian frequency profile on a per-token, per-axis, and per-band basis, while regularization anchors them to the analytic prior.
This gives the attention layer the two properties the failure analysis asks for: local variation, because geometry reliability changes across rays, boundaries, and materials; and task-adaptive selectivity, because different manipulation skills need different spatial frequencies.
Coarse tasks such as button pushing can rely mostly on stable low-frequency geometry, while precision tasks such as peg insertion or assembly may benefit from stronger dampening of unreliable high-frequency positional cues.
GM3A is therefore complementary to Geometric Cross-View Self-Supervision: cross-view supervision shapes what the shared features learn, while GM3A changes how uncertain coordinates are allowed to influence deployed attention.

\subsection{Training and Deployment}
\label{sec:method-train-deploy}

Training combines the base policy objective with the two \textsc{NISE} terms:
\begin{equation*}
  \mathcal{L}_{\text{total}} = \mathcal{L}_{\text{policy}} + \lambda_{\text{geo}}\,\mathcal{L}_{\text{geo}} + \lambda_{\text{reg}}\,\mathcal{L}_{\text{reg}}
\end{equation*}
where $\mathcal{L}_{\text{policy}}$ is the action-generation loss of the underlying policy class.
The cross-view geometry head contributes $\mathcal{L}_{\text{geo}}$ only during training.
GM3A contributes $\mathcal{L}_{\text{reg}}$ during training and remains part of the attention computation at inference.

At deployment, the cross-view geometry head is removed.
The deployed graph keeps the backbone, shared feature extractor, 3D attention, and action expert.
It does not require clean depth or a depth-completion preprocessor, and it adds no inference-time branch for cross-view reconstruction.

\section{Experiments}
\label{sec:experiments}

We evaluate \textsc{NISE} in both simulation and real-robot manipulation.
The experiments ask whether \textsc{NISE} can preserve clean-depth performance, improve robustness under realistic depth corruption beyond matched noisy training, transfer to real-world commodity RGB-D deployment, and attribute the gain to the proposed representation and attention components.
\subsection{Setup}
\label{sec:exp-setup}
\textbf{Simulation.} All comparisons use the same depth-corruption interface: the calibrated noise wrapper is applied at the dataloader before each policy consumes RGB-D.
This keeps the intervention at the sensor-observation level while preserving each baseline's original architecture and action decoder. We corrupt simulator depth with a calibrated procedural pipeline targeting three sensor-realistic failure modes: missing measurements, boundary-localized depth discontinuity errors, and plausible wrong-surface returns~\citep{handa2014simkinect,khoshelham2012kinect}. Parameter ranges are calibrated against a pre-characterization protocol on RealSense D405 and Azure Kinect capture of tabletop scenes. The main experiments use three corruption levels — clean, mild, and saturated — bracketing the calibrated RGB-D sensor operating ranges.
Saturated denotes the stress-test endpoint.

\begin{table}[t]
  \centering
  \caption{\textbf{Quantitative results on RLBench PerAct-18.}
  Per-task success rates (\%, mean $\pm$ standard deviation where available) on the standard 18-task PerAct suite under clean-depth evaluation.
  \textsc{NISE} reports mean $\pm$ standard deviation over repeated evaluations of a fixed checkpoint.
  Published baselines use clean-depth training, while \textsc{NISE} uses noisy-depth training.
  The best and second-best distinct means per column are bold and underlined, respectively, including ties.}
  \label{tab:peract-clean}
  \resizebox{\textwidth}{!}{%
  \footnotesize
  \setlength{\tabcolsep}{4pt}%
  \begin{tabular}{lccccccccc c}
    \toprule
    Method & Close Jar & Drag Stick & Insert Peg & Meat off Grill & Open Drawer & Place Cups & Place Wine & Push Buttons & Put Cupboard & \textbf{Avg.} \\
    \midrule
    PerAct~\citep{peract}                  & $55.2{\pm}4.7$ & $89.6{\pm}4.1$ & $5.6{\pm}4.1$  & $70.4{\pm}2.0$ & $88.0{\pm}5.7$ & $2.4{\pm}3.2$  & $44.8{\pm}7.8$ & $92.8{\pm}3.0$ & $28.0{\pm}4.4$ & $49.4$ \\
    RVT~\citep{rvt}                        & $52.0{\pm}2.5$ & $\underline{99.2{\pm}1.6}$ & $11.2{\pm}3.0$ & $88.0{\pm}2.5$ & $71.2{\pm}6.9$ & $4.0{\pm}2.5$  & $91.0{\pm}5.2$ & $\mathbf{100.0{\pm}0.0}$& $49.6{\pm}3.2$ & $62.9$ \\
    3D Diffuser Actor~\citep{tddda}        & $96.0{\pm}2.5$ & $\mathbf{100.0{\pm}0.0}$& $65.6{\pm}4.1$ & $\underline{96.8{\pm}1.6}$ & $89.6{\pm}4.1$ & $24.0{\pm}7.6$ & $93.6{\pm}4.8$ & $\underline{98.4{\pm}2.0}$ & $\mathbf{85.6{\pm}4.1}$ & $81.3$ \\
    RVT-2~\citep{rvt2}                     & $\mathbf{100.0{\pm}0.0}$& $99.0{\pm}1.7$ & $40.0{\pm}0.0$ & $\mathbf{99.0{\pm}1.7}$ & $74.0{\pm}11.8$& $38.0{\pm}4.5$ & $\underline{95.0{\pm}3.3}$ & $\mathbf{100.0{\pm}0.0}$& $66.0{\pm}4.5$ & $81.4$ \\
    ARP~\citep{arp}                        & $\underline{97.6}$         & $88.0$         & $53.2$         & $96.0$         & $\underline{90.4}$         & $48.0$         & $92.0$         & $\mathbf{100.0}$        & $68.0$         & $81.6$ \\
    TVVE~\citep{tvve}                      & $\mathbf{100.0{\pm}0.0}$& $\mathbf{100.0{\pm}0.0}$& $\mathbf{98.0{\pm}2.8}$ & $94.0{\pm}2.8$ & $90.0{\pm}2.8$ & $\underline{54.0{\pm}2.8}$ & $92.0{\pm}5.7$ & $\mathbf{100.0{\pm}0.0}$& $74.0{\pm}8.5$ & $\underline{86.6}$ \\
    \midrule
    \textbf{\textsc{NISE} (ours)}          & $\mathbf{100.0{\pm}0.0}$ & $\mathbf{100.0{\pm}0.0}$ & $\underline{92.0{\pm}3.5}$ & $96.0{\pm}2.0$ & $\mathbf{92.0{\pm}2.8}$ & $\mathbf{58.0{\pm}4.5}$ & $\mathbf{96.0{\pm}2.8}$ & $\mathbf{100.0{\pm}0.0}$ & $\underline{78.0{\pm}4.0}$ & $\mathbf{89.0}$ \\
    \midrule
    \midrule
    Method & Put Drawer & Put Safe & Screw Bulb & Slide Block & Sort Shape & Stack Blocks & Stack Cups & Sweep Dustpan & Turn Tap & \textbf{Avg.} \\
    \midrule
    PerAct~\citep{peract}                  & $51.2{\pm}4.7$ & $84.0{\pm}3.6$ & $17.6{\pm}2.0$ & $74.0{\pm}13.0$& $16.8{\pm}4.7$ & $26.4{\pm}3.2$ & $2.4{\pm}2.0$  & $52.0{\pm}0.0$ & $88.0{\pm}4.4$ & $49.4$ \\
    RVT~\citep{rvt}                        & $88.0{\pm}5.7$ & $91.2{\pm}3.0$ & $48.0{\pm}5.7$ & $81.6{\pm}5.4$ & $36.0{\pm}2.5$ & $28.8{\pm}3.9$ & $26.4{\pm}8.2$ & $72.0{\pm}0.0$ & $93.6{\pm}4.1$& $62.9$ \\
    3D Diffuser Actor~\citep{tddda}        & $96.0{\pm}3.6$ & $\mathbf{97.6{\pm}2.0}$ & $82.4{\pm}2.0$ & $97.6{\pm}3.2$ & $44.0{\pm}4.4$ & $68.3{\pm}3.3$ & $47.2{\pm}8.5$ & $84.0{\pm}4.4$ & $\underline{99.2{\pm}1.6}$ & $81.3$ \\
    RVT-2~\citep{rvt2}                     & $96.0{\pm}0.0$ & $\underline{96.0{\pm}2.8}$ & $\underline{88.0{\pm}4.9}$ & $92.0{\pm}2.8$ & $35.0{\pm}7.1$ & $\mathbf{80.0{\pm}2.8}$ & $69.0{\pm}5.9$ & $\mathbf{100.0{\pm}0.0}$& $99.0{\pm}1.7$& $81.4$ \\
    ARP~\citep{arp}                        & $\underline{99.2}$         & $94.4$         & $85.6$         & $\underline{98.4}$         & $35.2$         & $55.2$         & $\mathbf{76.8}$         & $90.4$         & $\mathbf{100.0}$        & $81.6$ \\
    TVVE~\citep{tvve}                      & $\mathbf{100.0{\pm}0.0}$& $78.0{\pm}2.8$ & $86.0{\pm}2.8$ & $\mathbf{100.0{\pm}0.0}$& $\underline{62.0{\pm}8.5}$ & $74.0{\pm}2.8$ & $64.0{\pm}5.7$ & $92.0{\pm}5.7$ & $\mathbf{100.0{\pm}0.0}$& $\underline{86.6}$ \\
    \midrule
    \textbf{\textsc{NISE} (ours)}          & $\mathbf{100.0{\pm}0.0}$ & $90.0{\pm}3.5$ & $\mathbf{90.0{\pm}2.8}$ & $\mathbf{100.0{\pm}0.0}$ & $\mathbf{66.0{\pm}5.5}$ & $\underline{78.0{\pm}3.5}$ & $\underline{70.0{\pm}4.0}$ & $\underline{96.0{\pm}2.8}$ & $\mathbf{100.0{\pm}0.0}$ & $\mathbf{89.0}$ \\
    \bottomrule
  \end{tabular}}
\end{table}

For \Cref{tab:noisy-summary}, all methods on RLBench and MetaWorld are trained with clean, mild, and saturated depth sampled with equal probability.
We provide a cross-benchmark simulation check on two benchmarks: RLBench PerAct-18~\citep{rlbench,peract} and MetaWorld~\citep{metaworld}, under both clean depth (\Cref{tab:peract-clean}) and realistic depth noise (\Cref{tab:noisy-summary}).
For \textsc{NISE}, RLBench uses cross-view depth supervision, while single-view MetaWorld uses clean simulator depth as the auxiliary geometry target during training (Appendix~\ref{app:training-configuration}).
We use $25$ episodes per condition and match episode counts within each cell.
We report mean success over three independent training runs in \Cref{tab:noisy-summary}.

\begin{wrapfigure}{r}{0.50\linewidth}
  \centering
  \includegraphics[width=\linewidth]{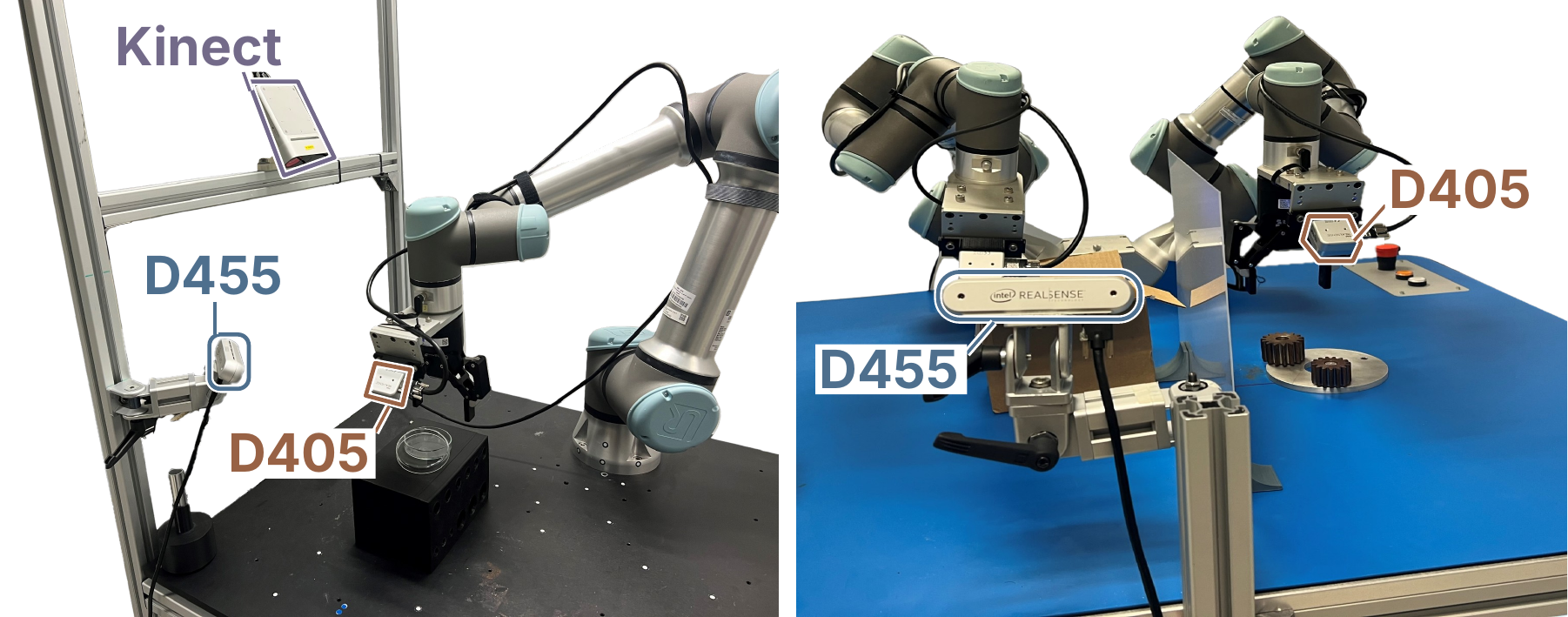}
  \caption{Real-robot setup: single-arm UR5e (left) and bimanual UR3e (right). Camera labels identify Azure Kinect and RealSense D455/D405.}
  \label{fig:real-robot-setup}
\end{wrapfigure}

\textbf{Real-Robot.}
We evaluate on three single-arm and two bimanual tasks involving transparent obstacles, specular surfaces, transparent manipulation targets, and thin or reflective parts, so depth failures arise from the task geometry naturally rather than from an isolated sensor diagnostic.
The single-arm and bimanual setups are shown in \Cref{fig:real-robot-setup}.

\Needspace{16\baselineskip}
\textbf{Implementation.}
Each baseline retains its native action expert.
\textsc{NISE} and the 3DDA-family rows use a DDPM-based diffusion action expert, while RVT-2 uses its native keypose predictor.
For GCVSS, invalid source depth is discarded before warping, splatted pixels are z-buffered in the target view, and supervision is applied only where the projected geometry is valid and visible.
For GM3A, two lightweight per-token MLPs predict residuals around an analytic attenuation prior initialized from calibrated sensor-scale depth variance.
The resulting coefficient $\alpha=\alpha^{\text{prior}}+\delta$ is regularized toward the prior through $\mathcal{L}_{\text{reg}}$.
We use VGGT-$\Omega$~\citep{vggt-omega} as the frozen visual backbone.
Unless otherwise noted, all experiments use $\lambda_{\text{geo}}=0.1$ and $\lambda_{\text{reg}}=1.0$, selected by validation sweep.
\subsection{Results}
\label{sec:exp-rlbench}
\textbf{Simulation.} The RLBench experiments ask whether noisy-depth training alone is sufficient to prevent the depth-noise cliff.
\Cref{tab:peract-clean} first checks clean-depth inheritance using the standard PerAct-18.
\textsc{NISE} matches or slightly exceeds the strongest clean-depth baselines, so the noise-robust representation does not have to be purchased by sacrificing the nominal benchmark setting.
As an image-based reference, PerAct's single-view RGB-D Image-BC baselines achieve mean success rates of $1.3\%$ (CNN) and $0.9\%$ (ViT) across the 18 tasks with 100 demonstrations per task under noiseless evaluation~\citep{peract}.
These results motivate exploiting explicit 3D structure for precise manipulation.
\textsc{NISE} addresses the accompanying challenge: maintaining reliable spatial representations when the depth measurements used to construct them are corrupted.
The noisy-depth comparison is summarized visually in \Cref{fig:teaser}b and numerically in \Cref{tab:noisy-summary}.

\begin{table}[t]
  \centering
  \caption{\textbf{Robustness to depth corruption after noise-aware training.}
  Task-averaged success rates (\%) on RLBench PerAct-18 and MetaWorld under clean, mild-noise, and saturated-noise evaluation.
  Values are means $\pm$ standard deviations across three independent training runs.}
  \label{tab:noisy-summary}
  \footnotesize
  \setlength{\tabcolsep}{2pt}
  \begin{minipage}[t]{0.50\linewidth}
    \centering
    \begin{tabular}{@{}lccc@{}}
      \toprule
      \multicolumn{4}{@{}l@{}}{\textbf{RLBench PerAct-18}} \\
      \midrule
      Method & Clean & Mild & Saturated \\
      \midrule
      3DDA~\citep{tddda}       & $64.8{\pm}9.1$ & $60.2{\pm}9.7$ & $47.8{\pm}12.0$ \\
      RVT-2~\citep{rvt2}       & $65.2{\pm}9.5$ & $62.4{\pm}9.4$ & $48.5{\pm}13.8$ \\
      SAM2Act~\citep{sam2act}  & $68.1{\pm}9.8$ & $63.7{\pm}8.9$ & $50.9{\pm}11.9$ \\
      \textbf{\textsc{NISE}}   & $\mathbf{89.0{\pm}4.3}$ & $\mathbf{85.7{\pm}4.5}$ & $\mathbf{84.1{\pm}4.6}$ \\
      \bottomrule
    \end{tabular}
  \end{minipage}\hfill%
  \begin{minipage}[t]{0.49\linewidth}
    \centering
    \begin{tabular}{@{}lccc@{}}
      \toprule
      \multicolumn{4}{@{}l@{}}{\textbf{MetaWorld}} \\
      \midrule
      Method & Clean & Mild & Saturated \\
      \midrule
      DP3~\citep{dp3}             & $68.3{\pm}13.6$ & $60.5{\pm}14.2$ & $36.8{\pm}19.8$ \\
      FlowPolicy~\citep{flowpolicy}
      & $69.6{\pm}13.8$ & $61.4{\pm}13.9$ & $37.2{\pm}19.3$ \\
      ManiFlow~\citep{maniflow}   & $70.5{\pm}15.7$ & $62.8{\pm}15.1$ & $37.0{\pm}18.5$ \\
      \textbf{\textsc{NISE}}       & $\mathbf{79.0{\pm}6.3}$ & $\mathbf{78.2{\pm}8.6}$ & $\mathbf{77.4{\pm}9.4}$ \\
      \bottomrule
    \end{tabular}
  \end{minipage}
\end{table}

\begin{figure}[t]
  \centering
  \includegraphics[width=0.98\linewidth]{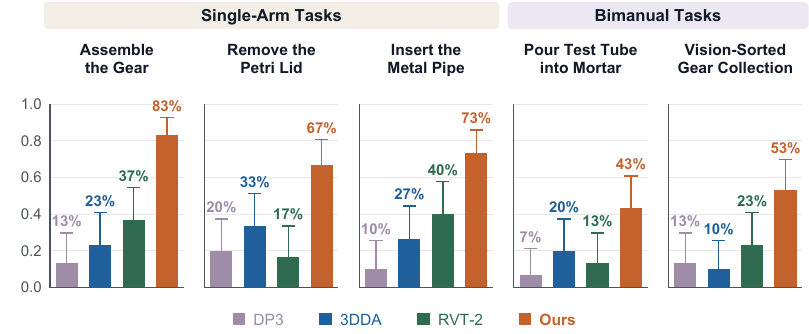}
  \caption{Real-robot Success Rate. Each task group compares DP3, 3DDA, RVT-2, and \textsc{NISE}, with $N=30$ trials per task and method.}
  \label{fig:real-robot}
\end{figure}

\textbf{Real-Robot.}
The real-robot validation asks whether the learned spatial encoding remains useful under genuine commodity-sensor noise, where clean depth is not available. \Cref{fig:real-robot} summarizes the real-robot results as grouped per-task bars comparing DP3, 3DDA, RVT-2, and \textsc{NISE}.
The gap observed in simulation transfers to deployment.
The CDM $\times$ policy control in \Cref{app:cdm-policy-realrobot} examines the complementary effects of depth repair and \textsc{NISE} on real-robot lid removal.

\subsection{Ablations}
\label{sec:exp-ablations}
\textbf{Per-component ablation.}
Table~\ref{tab:ablation} separates two effects that should not be conflated: the stronger frozen-ViT+DPT action-expert stack, and the two proposed mechanisms.
The frozen-ViT+DPT control already outperforms the best matched-noise baseline in \Cref{tab:noisy-summary}, so we report it explicitly rather than attributing that gain to \textsc{NISE}.
Under noisy-depth training, GCVSS and GM3A individually improve success from $75.5\%$ to $80.5\%$ and $78.2\%$, respectively, while their combination reaches $85.7\%$.
The combination also achieves the highest mean under clean-depth training ($68.6\%$).
The stronger combined performance suggests synergy between learning geometry-consistent features through cross-view supervision and adapting positional attention to coordinate uncertainty.

\begin{table}[ht]
  \centering
  \caption{\textbf{Baseline control and component ablation.}
  Mean success rate on RLBench PerAct-18 under \textit{mild-noise depth evaluation}. Both clean-depth and noisy-depth blocks use the same frozen-ViT+DPT encoder and action expert, isolating the effects of GCVSS and GM3A. $\Delta$FLOPs reports the additional deployed inference cost in GFLOPs.}
  \label{tab:ablation}
  \scriptsize
  \setlength{\tabcolsep}{1.8pt}
  \begin{tabular}{@{}c@{\hspace{0.015\linewidth}}c@{}}
    \begin{minipage}[t]{0.49\linewidth}
      \centering
      \textbf{Trained on Clean Depth}\\[-0.5mm]
      \begin{tabular}{@{}lcccc@{}}
        \toprule
        Setting & GCVSS & GM3A & SR (\%) & $\Delta$FLOPs \\
        \midrule
        Baseline & $\times$ & $\times$ & $52.1{\pm}6.1$ & $+0.0$ \\
        + GM3A & $\times$ & $\checkmark$ & $53.3{\pm}5.8$ & $+0.8$ \\
        + GCVSS & $\checkmark$ & $\times$ & $62.0{\pm}5.2$ & $+0.0$ \\
        + Both & $\checkmark$ & $\checkmark$ & $68.6{\pm}4.8$ & $+0.8$ \\
        \bottomrule
      \end{tabular}
    \end{minipage}
    &
    \begin{minipage}[t]{0.49\linewidth}
      \centering
      \textbf{Trained on Noisy Depth}\\[-0.5mm]
      \begin{tabular}{@{}lcccc@{}}
        \toprule
        Setting & GCVSS & GM3A & SR (\%) & $\Delta$FLOPs \\
        \midrule
        Baseline & $\times$ & $\times$ & $75.5{\pm}4.5$ & $+0.0$ \\
        + GM3A & $\times$ & $\checkmark$ & $78.2{\pm}4.0$ & $+0.8$ \\
        + GCVSS & $\checkmark$ & $\times$ & $80.5{\pm}3.5$ & $+0.0$ \\
        \textbf{\textsc{NISE}} & $\checkmark$ & $\checkmark$ & $\mathbf{85.7{\pm}4.5}$ & $+0.8$ \\
        \bottomrule
      \end{tabular}
    \end{minipage}
  \end{tabular}
\end{table}
Additional details and results are provided in the appendix, including the full corruption equations and D405/Azure Kinect calibration protocol; implementation, runtime, and training details; per-task MetaWorld results and precision-task GM3A analyses; feature-invariance and attention-dampening diagnostics, including the multi-view-agreement failure on transparent objects; and ablations over adaptive attenuation and loss weights.
The appendix also reports backbone-matched ablations that insert GCVSS and GM3A into native 3DDA and RVT-2, isolating mechanism gains from the VGGT-$\Omega$ backbone; \textsc{NISE} combined with depth-repair front ends~\citep{wei2024d3roma,marigold_dc}; and the full real-robot protocol with per-task trial counts and confidence intervals.

\FloatBarrier{}

\section{Limitations and Conclusion}
\label{sec:discussion}
\textbf{Limitations.}
The current instantiation assumes calibrated multi-view RGB-D, so cross-view geometry supervision requires sufficient camera overlap and calibration quality.
It can also fail when multiple views agree on the same wrong surface, such as transparent objects for which different cameras recover a view-consistent background depth rather than the true object geometry. GCVSS helps when view-specific artifacts disagree, but can reinforce systematic wrong surfaces when all cameras observe the same background through a transparent object. GM3A partially mitigates this by dampening high-frequency position cues, but does not fully solve systematic multi-view depth hallucination.
Depth discontinuities and wrong-surface returns can produce biased or multimodal coordinate errors that a local Gaussian prior does not fully capture.
Mixture models, ray-aligned covariance, and non-parametric uncertainty representations are promising extensions for modeling this richer error structure.
These are scope boundaries rather than deployment blockers: \textsc{NISE} does not require clean depth at inference, and its supervision can be built from robot-collected multi-view data.

\textbf{Conclusion.}
Realistic depth corruption exposes a missing capability in current 3D manipulation policies: they lift depth into geometry, but their spatial encoders often treat corrupted measurements as reliable.
We characterize the depth-noise cliff, show that matched noisy training only partially recovers performance, and introduce \textsc{NISE} to learn noise-invariant spatial encodings through cross-view geometry supervision and uncertainty-aware 3D attention.
Across simulation, ablations, and real-robot tasks, the result supports a simple design principle: 3D policies for commodity RGB-D sensors should learn geometry features that remain trustworthy when depth is imperfect.

\acknowledgments{
This research is supported by National Robotics Programme (NRP) 2.0 funding initiative ``Domain-specific Robotics Foundation Models for Manufacturing (DS-RFM)''.
}

\bibliography{references}

\clearpage
\appendix
\newcommand{\appgraphics}[2]{\includegraphics[#2]{#1}}

\FloatBarrier{}
\section{Depth Corruption Model}
\label{app:noise-model}

We apply the same procedural depth wrapper to RLBench and MetaWorld before policy input construction.
The wrapper combines depth-dependent axial noise, lateral displacement near discontinuities, and invalid or wrong-surface measurements.
Schematically,

\begin{equation}
  \tilde D(\mathbf{u})
  =
  M(\mathbf{u})
  \left[
    D(\mathbf{u})
    + \epsilon_{\mathrm{range}}(\mathbf{u})
    + \epsilon_{\mathrm{edge}}(\mathbf{u})
    + \epsilon_{\mathrm{surface}}(\mathbf{u})
  \right],
  \label{eq:app-compact-corruption}
\end{equation}

where $M$ removes invalid measurements and the three error terms denote range, boundary, and wrong-surface errors.
Wrong-surface returns arise from displaced depth lookup in \Cref{eq:app-lateral,eq:app-observed-depth}, rather than an additional independent Bernoulli process.
The corruption types are activated jointly because these failures commonly co-occur near object boundaries and transparent surfaces.

For calibration, we captured repeated frames of static tabletop scenes containing planar boards and manipulation objects with the D405 and Azure Kinect after intrinsic and extrinsic calibration.
We estimated invalid-pixel rates, disagreement near depth discontinuities, and axial residuals as a function of range.
The mild and saturated presets bracket these observed statistics and remain fixed across policies and benchmarks.

Let $D$ denote the clean simulator depth map at pixel $\mathbf{u}={[u,v]}^\top$, with intrinsics $K$, focal length $f$, and stereo baseline $b$.
For stereo-style axial noise, we sample disparity noise and invert back to metric depth:
\begin{equation}
  s(\mathbf{u}) = \frac{f b}{D(\mathbf{u})}, \qquad
  \tilde s(\mathbf{u}) = s(\mathbf{u}) + \eta_s(\mathbf{u}), \qquad
  D_{\mathrm{ax}}(\mathbf{u}) = \frac{f b}{\tilde s(\mathbf{u})},
  \label{eq:app-axial-noise}
\end{equation}
where $\eta_s \sim \mathcal{N}(0,\sigma_s^2)$.
For small perturbations, this gives the familiar depth-squared growth
\begin{equation}
  \sigma_z(z) \approx \frac{z^2}{f b}\,\sigma_s .
  \label{eq:app-z2}
\end{equation}

\begin{figure}[htbp]
  \centering
  \begin{subfigure}[t]{0.23\linewidth}
    \centering
    \appgraphics{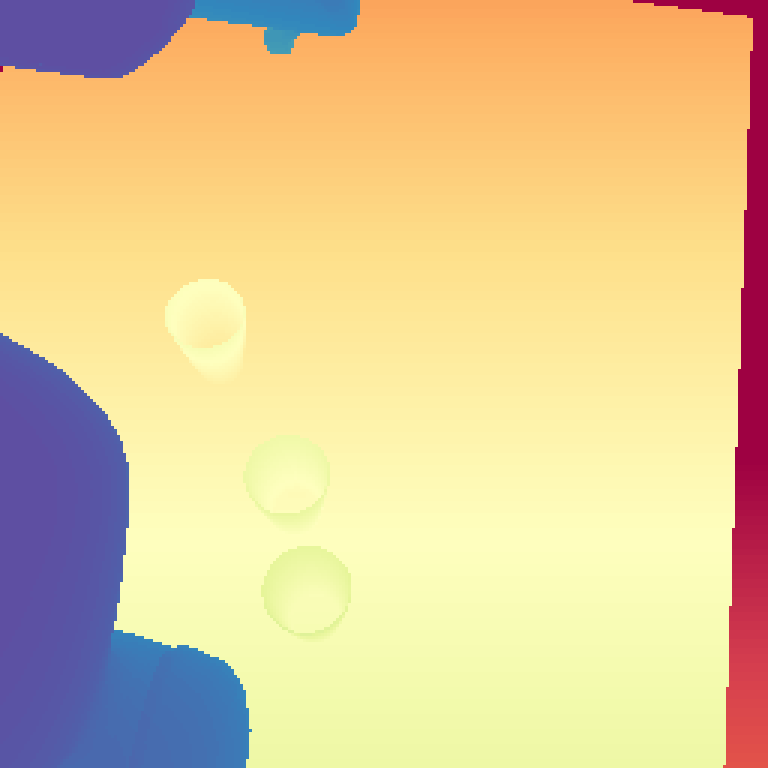}{width=\linewidth}
    \caption{Sim clean}
  \end{subfigure}\hfill
  \begin{subfigure}[t]{0.23\linewidth}
    \centering
    \appgraphics{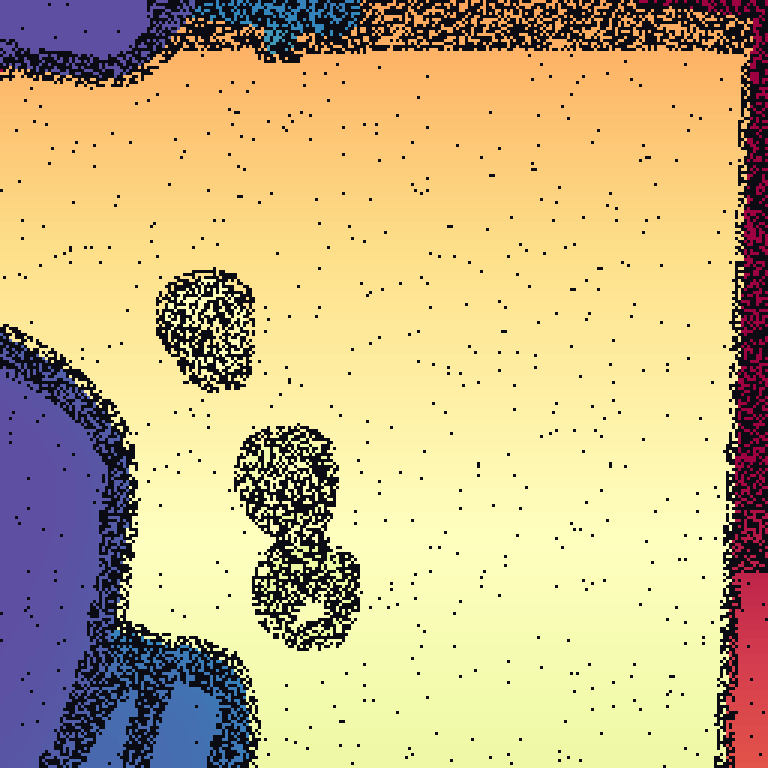}{width=\linewidth}
    \caption{Sim holes}
  \end{subfigure}\hfill
  \begin{subfigure}[t]{0.23\linewidth}
    \centering
    \appgraphics{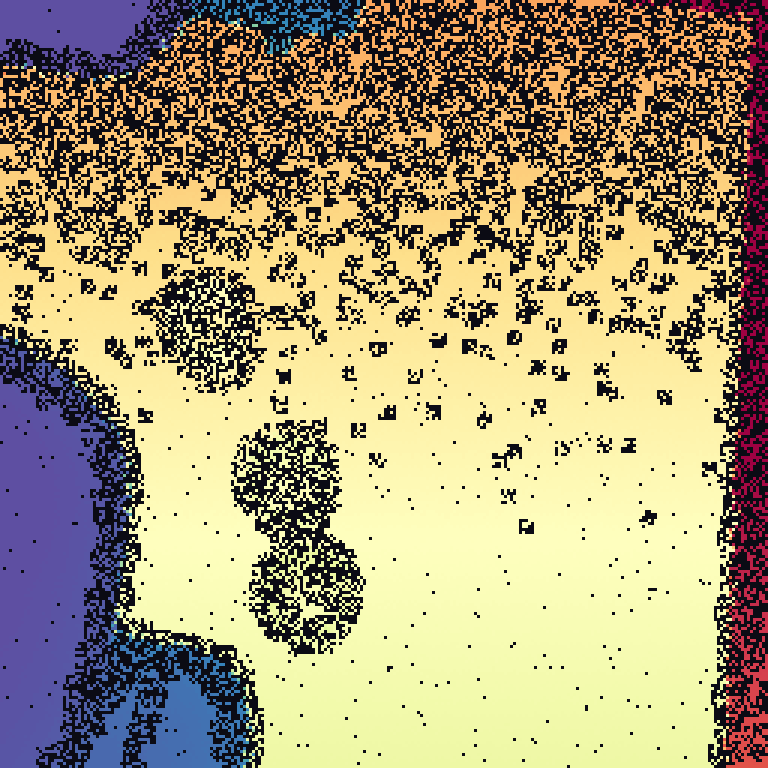}{width=\linewidth}
    \caption{Sim boundary}
  \end{subfigure}\hfill
  \begin{subfigure}[t]{0.23\linewidth}
    \centering
    \appgraphics{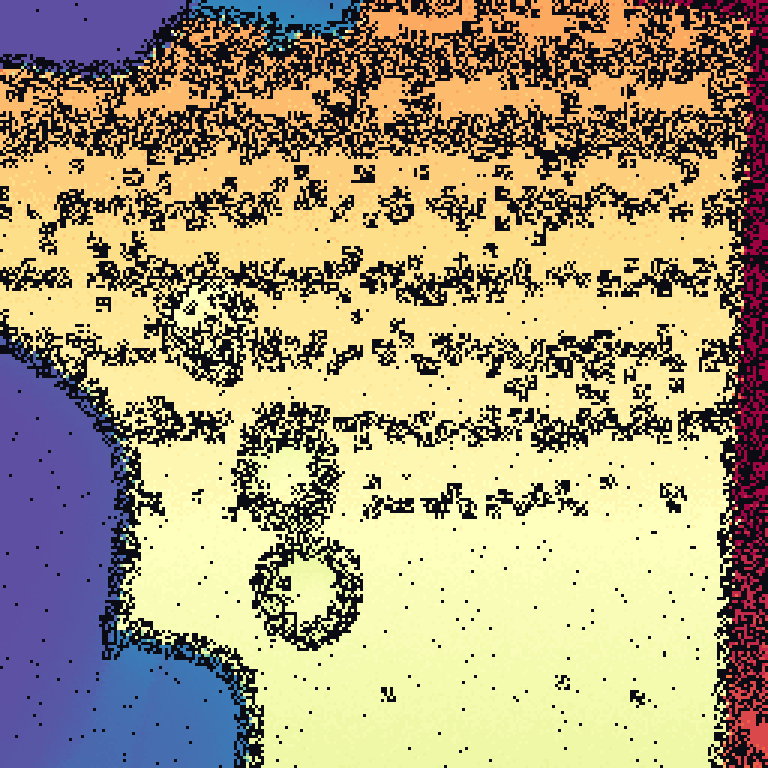}{width=\linewidth}
    \caption{Sim displaced}
  \end{subfigure}
  \caption{\textbf{Depth-corruption families used as a controlled diagnostic.}
  The panels show the simulator depth wrapper applied to a clean rendered frame, isolating random holes, boundary-localized artifacts, and displaced lookup near discontinuities.
  The figure is a compact visual explanation of the procedural knobs in \Cref{eq:app-lateral,eq:app-dropout,eq:app-observed-depth}; sensor-specific calibration is captured by the preset scales rather than by exact rendering of a single camera model.}
  \label{fig:app-depth-corruption-families}
\end{figure}

Lateral jitter is applied in pixel space before lookup:
\begin{equation}
  \tilde{\mathbf{u}} = \mathbf{u} + \boldsymbol{\eta}_{\ell},
  \qquad
  \boldsymbol{\eta}_{\ell} \sim \mathcal{N}(\mathbf{0},\sigma_{\ell}^{2} I_2).
  \label{eq:app-lateral}
\end{equation}
The invalid-depth mask combines sensor-valid depth, edge dropout, and random holes:
\begin{equation}
  M(\mathbf{u}) =
  M_{\mathrm{valid}}(\mathbf{u})\,
  M_{\mathrm{edge}}(\mathbf{u})\,
  M_{\mathrm{hole}}(\mathbf{u}),
  \label{eq:app-mask}
\end{equation}
with
\begin{equation}
  M_{\mathrm{edge}}(\mathbf{u}) \sim
  \operatorname{Bernoulli}\!\left(1 - p_{\mathrm{edge}}\,
  \mathbf{1}\{\|\nabla D(\mathbf{u})\| > \tau\}\right),
  \qquad
  M_{\mathrm{hole}}(\mathbf{u}) \sim
  \operatorname{Bernoulli}(1 - p_{\mathrm{hole}}).
  \label{eq:app-dropout}
\end{equation}
The observed depth is then
\begin{equation}
  \tilde D(\mathbf{u}) =
  M(\mathbf{u})\,
  D_{\mathrm{ax}}\!\left(\tilde{\mathbf{u}}\right),
  \label{eq:app-observed-depth}
\end{equation}
where zero-valued pixels are treated as invalid by downstream point-cloud and token-lifting code.
The model follows the analytic sensor-noise lineage used in depth-sensor simulation and characterization~\citep{handa2014simkinect,khoshelham2012kinect}.
\Cref{fig:app-depth-corruption-families} shows the corresponding simulator-side visual diagnostic.

\begin{table}[htbp]
  \centering
  \caption{\textbf{Depth-noise presets.}
  The main paper uses clean, mild, and saturated; saturated corresponds to the high-stress endpoint used in the teaser and main summary table.
  Saturated is named \texttt{moderate} in the implementation, with the same parameter values.}
  \label{tab:noise-presets}
  \small
  \setlength{\tabcolsep}{6pt}
  \begin{tabular}{@{}lcccc@{}}
    \toprule
    Preset & $\sigma_{\ell}$ (px) & $\sigma_s$ (px) & $p_{\mathrm{edge}}$ & $p_{\mathrm{hole}}$ \\
    \midrule
    Clean & 0.00 & 0.000 & 0.000 & 0.0000 \\
    Mild & 0.30 & 0.150 & 0.300 & 0.0020 \\
    Saturated & 0.50 & 0.250 & 0.500 & 0.0050 \\
    \bottomrule
  \end{tabular}
\end{table}

\FloatBarrier{}
\section{Visual Backbone and Policy Ablations}
\label{app:visual-backbone}

\subsection{Visual Backbone}

Our policy uses the public \texttt{facebook/VGGT-Omega} checkpoint as a frozen visual backbone~\citep{vggt-omega,vggt}, together with a trainable DPT aggregator and a DDPM action expert.
The backbone provides geometry-pretrained features, while GCVSS and GM3A shape their use for manipulation under depth corruption.
We evaluate the contribution of these mechanisms through the same-stack controls and native-policy ablations below.

\subsection{Same-Stack Base-Model Control}
\label{app:same-stack-base-control}

\begin{figure}[!htbp]
  \centering
  \begin{minipage}[c]{0.40\linewidth}
    \centering
    \includegraphics[width=\linewidth]{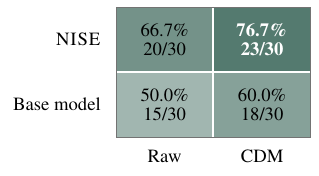}
  \end{minipage}\hfill
  \begin{minipage}[c]{0.56\linewidth}
    \caption{\textbf{CDM $\times$ policy control on real-robot lid removal.}
    Evaluation uses Remove the Petri Lid under real sensor noise, with $30$ trials per configuration.
    Each cell reports the success rate and successful trials out of $30$.
    The base model uses the same policy stack as \textsc{NISE} with GCVSS and GM3A disabled.}
    \label{fig:cdm-policy-realrobot}
  \end{minipage}
\end{figure}

Clean-depth evaluation tests whether the proposed mechanisms preserve useful spatial information when the input geometry is reliable.
We fix the frozen VGGT-$\Omega$ backbone, DPT aggregator, and DDPM action expert, and toggle GCVSS and GM3A together.
Both variants use the same training recipe.
Under clean-depth evaluation on RLBench PerAct-18, the base model with both modules disabled achieves $79.0{\pm}5.0\%$ success, while enabling both modules yields $89.0{\pm}4.3\%$ for \textsc{NISE} (mean $\pm$ standard deviation).
The gain suggests that the learned encoding also benefits control with reliable depth, extending its usefulness beyond corrupted inputs.
Only this within-stack improvement is attributed to the proposed mechanisms.
The main component ablation (\Cref{tab:ablation}) evaluates the same stack under mild noise.
\Cref{fig:cdm-policy-realrobot} provides the corresponding real-robot base-model control.

\subsection{Native-Policy Ablations on 3DDA and RVT-2}
\label{app:native-insertions}

We next test whether the mechanisms remain useful when the visual backbone and action architecture differ from our final stack.
\Cref{tab:native-policy-insertions} inserts GCVSS and GM3A into native 3DDA and RVT-2 on push\_buttons, insert\_peg, and stack\_cups.
The 3DDA block retains its CLIP-RN50/ResNet-style feature pyramid and diffusion action expert; RVT-2 retains its virtual-view/keypose architecture.

\begin{table}[htbp]
  \centering
  \caption{\textbf{Native-stack ablations on RLBench.}
  Success rates (\%) are means over three RLBench tasks: push\_buttons, insert\_peg, and stack\_cups.
  The 3DDA block keeps the native 3DDA feature pyramid and diffusion action expert; the RVT-2 block keeps the virtual-view/keypose policy family.
  Within each policy block, all four variants are fine-tuned on saturated-noise demonstrations using the same protocol, varying only the inclusion of GCVSS and GM3A.}
  \label{tab:native-policy-insertions}
  \small
  \setlength{\tabcolsep}{5pt}
  \begin{tabular}{@{}llccrrr@{}}
    \toprule
    Policy stack & Ablation row & GCVSS & GM3A & Clean & Mild & Saturated \\
    \midrule
    \multicolumn{7}{@{}l}{\textit{Native 3DDA insertion}} \\
    3DDA & Plain noise-aware & $\times$ & $\times$ & 41.0 & 54.0 & 56.7 \\
    3DDA & + GCVSS & $\checkmark$ & $\times$ & 45.0 & 62.0 & 63.5 \\
    3DDA & + GM3A & $\times$ & $\checkmark$ & 43.0 & 59.0 & 61.0 \\
    3DDA & + GCVSS + GM3A & $\checkmark$ & $\checkmark$ & 47.0 & 68.0 & 70.3 \\
    \midrule
    \multicolumn{7}{@{}l}{\textit{Native RVT-2 insertion}} \\
    RVT-2 & Plain noise-aware & $\times$ & $\times$ & 52.0 & 34.0 & 24.0 \\
    RVT-2 & + GCVSS & $\checkmark$ & $\times$ & 55.0 & 42.0 & 35.0 \\
    RVT-2 & + GM3A & $\times$ & $\checkmark$ & 54.0 & 40.0 & 34.0 \\
    RVT-2 & + GCVSS + GM3A & $\checkmark$ & $\checkmark$ & 57.0 & 48.0 & 44.0 \\
    \bottomrule
  \end{tabular}
\end{table}

The 3DDA variants' higher success under noisy evaluation is consistent with specialization to their saturated-noise training distribution.
The combined modules give the strongest noisy-depth results in both policy families, raising saturated success from $56.7\%$ to $70.3\%$ in 3DDA and from $24.0\%$ to $44.0\%$ in RVT-2.
The consistent direction of improvement supports using geometry supervision and adaptive positional attention across these two architectures, while the different absolute success rates show that the underlying policy stack still matters.

\FloatBarrier{}
\section{Depth-Repair Front-End Comparisons}
\label{app:depth-repair}

Depth repair changes the observations supplied to a policy, while \textsc{NISE} changes how the policy encodes those observations.
We test whether repair remains useful after the policy has learned to tolerate depth errors.
\Cref{tab:depth-repair-comparison} compares \textsc{NISE} with raw depth and with D3RoMa~\citep{wei2024d3roma} or Marigold-DC~\citep{marigold_dc} preprocessing.
Both front ends yield modestly higher noisy-depth means while retaining similar clean-input performance.
The main experiments use raw depth without a separate repair model.

\begin{table}[htbp]
  \centering
  \caption{\textbf{Effect of depth repair on \textsc{NISE}.}
  Mean success rate $\pm$ standard deviation (\%) on RLBench PerAct-18 under clean, mild-noise, and saturated-noise evaluation.
  We compare \textsc{NISE} with raw depth and with depth preprocessed by D3RoMa or Marigold-DC before the policy consumes RGB-D.
  Best means per column are bold.}
  \label{tab:depth-repair-comparison}
  \small
  \setlength{\tabcolsep}{4pt}
  \begin{tabular}{@{}p{0.32\linewidth}ccc@{}}
    \toprule
    Input / policy & Clean & Mild & Saturated \\
    \midrule
    \textbf{\textsc{NISE}} + raw depth & $89.0{\pm}4.3$ & $85.7{\pm}4.5$ & $84.1{\pm}4.6$ \\
    \textbf{\textsc{NISE}} + D3RoMa repair & $89.0{\pm}4.2$ & $86.1{\pm}4.3$ & $84.8{\pm}4.5$ \\
    \textbf{\textsc{NISE}} + Marigold-DC repair & $\mathbf{89.1{\pm}4.1}$ & $\mathbf{86.4{\pm}4.2}$ & $\mathbf{85.2{\pm}4.4}$ \\
    \bottomrule
  \end{tabular}
\end{table}

\Needspace{5\baselineskip}
\subsection{CDM and NISE on Real-Robot}
\label{app:cdm-policy-realrobot}

The real-robot control separates the effect of repairing the depth input from that of changing the policy's spatial encoding.
We evaluate the same-stack base model and \textsc{NISE} with raw or CDM-repaired depth~\citep{liu2025cdm} on Remove the Petri Lid.
Both policies share the visual backbone and action expert; \textsc{NISE} additionally uses GCVSS and GM3A.

CDM adds $10.0$ percentage points for both policies, and \textsc{NISE} adds $16.7$ percentage points with either depth input.
The retained gains support complementarity on this task.

\FloatBarrier{}
\section{GCVSS Target-Construction Ablation}
\label{app:gcvss-ablation}

\begin{figure}[htbp]
  \centering
  \begin{subfigure}[t]{0.31\linewidth}
    \centering
    \appgraphics{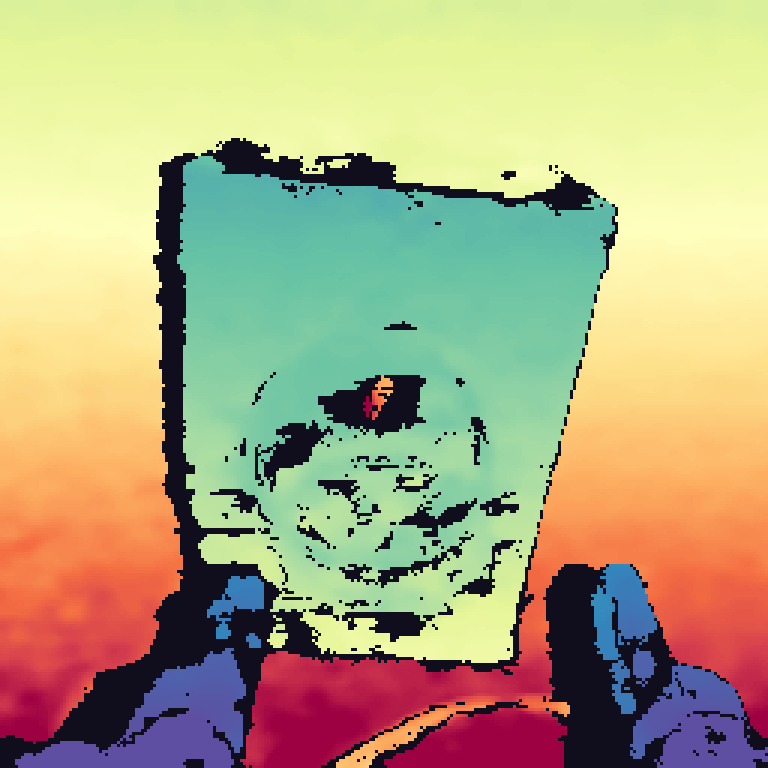}{width=\linewidth}
    \caption{Target-view depth}
  \end{subfigure}\hfill
  \begin{subfigure}[t]{0.31\linewidth}
    \centering
    \appgraphics{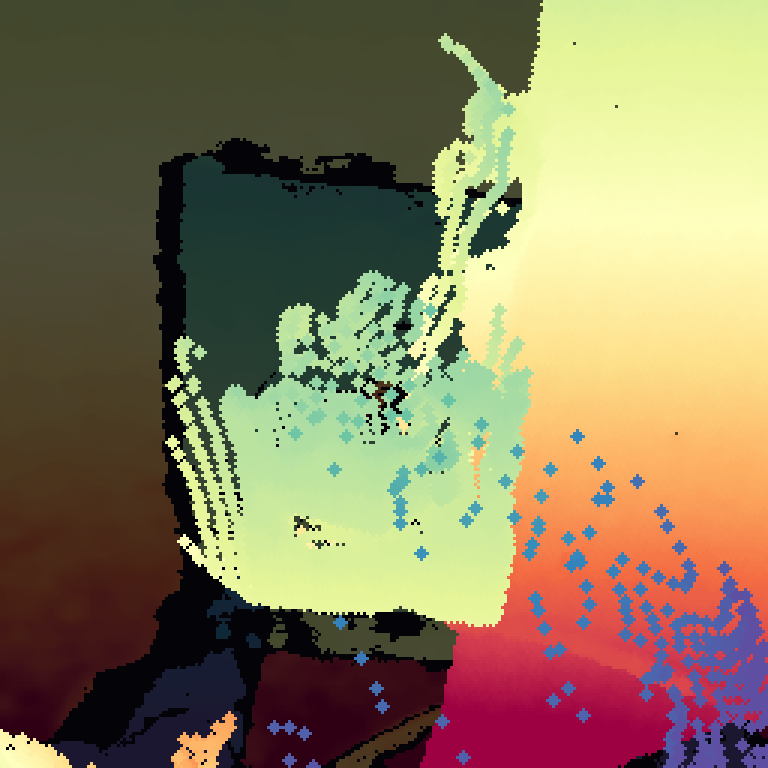}{width=\linewidth}
    \caption{Source depth warped to target}
  \end{subfigure}\hfill
  \begin{subfigure}[t]{0.31\linewidth}
    \centering
    \appgraphics{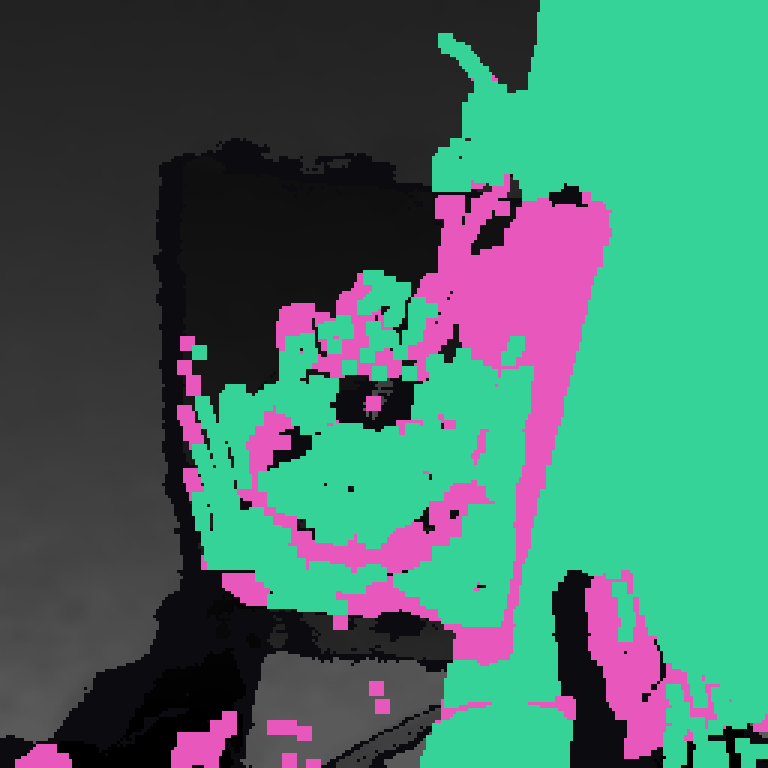}{width=\linewidth}
    \caption{Valid agreement mask}
  \end{subfigure}
  \caption{\textbf{Cross-view depth becomes a geometry target only after calibrated warping.}
  The source camera is forward-warped into the target camera frame and filtered by a visibility/validity mask before it supervises the shared feature map.
  This is the practical difference between a raw cross-camera noisy target and GCVSS: the auxiliary signal is still noisy, but it is expressed in the same camera frame as the target feature map.}
  \label{fig:app-gcvss-warp}
\end{figure}

Cross-camera depth can supervise geometry only if the target describes the surface visible in the receiving camera.
Directly pairing depth maps from different viewpoints mixes changes in surface geometry with camera pose and occlusion.
GCVSS forward-warps source depth into the target camera frame and supervises only valid, visible splats (\Cref{fig:app-gcvss-warp}).
\Cref{tab:gcvss-target-construction} isolates the contributions of camera-frame alignment, z-buffer visibility filtering, and validity masking under the same policy stack and training budget.

\begin{table}[htbp]
  \centering
  \caption{\textbf{GCVSS target-construction and component ablation.}
  Success rates (\%) on RLBench Stack Cups under saturated depth corruption.
  All variants use the same policy stack and training budget.
  The component controls remove z-buffer visibility filtering or validity masking from the full calibrated cross-view target construction.
  Clean-depth targets are simulation references; the best noisy-depth variant is bold.}
  \label{tab:gcvss-target-construction}
  \small
  \setlength{\tabcolsep}{8pt}
  \begin{tabular}{@{}lr@{}}
    \toprule
    Target construction & Success (\%) \\
    \midrule
    \multicolumn{2}{@{}l}{\textit{Clean-depth references}} \\
    Per-camera clean target & 64.0 \\
    Masked clean consensus & 62.4 \\
    \midrule
    \multicolumn{2}{@{}l}{\textit{Noisy-depth targets}} \\
    Raw cross-camera target & 38.2 \\
    Warped target without z-buffer visibility & 55.2 \\
    Warped target without validity mask & 58.4 \\
    \textbf{Full GCVSS} & \textbf{61.6} \\
    \bottomrule
  \end{tabular}
\end{table}

Full GCVSS reaches $61.6\%$ success, compared with $38.2\%$ for raw cross-camera targets.
Removing z-buffer filtering or the validity mask reduces success by $6.4$ or $3.2$ percentage points, respectively.
These differences show why spatial alignment alone is insufficient: the target must also select valid measurements from the corresponding visible surface.
The noisy cross-view target also approaches the per-camera clean target ($64.0\%$) and masked clean consensus ($62.4\%$).
This supports using aligned noisy observations as geometry supervision when clean depth is unavailable.

The accompanying training diagnostics help explain the effect of alignment.
Mean cross-camera depth correlation on the sampled training data rises from Pearson $r=0.14$ for raw targets to $0.367$ after warping.
At the final logged diagnostic step, the cosine between auxiliary and policy gradients changes from $-0.050$ to $+0.027$, close to the clean-target reference ($+0.024$).
These small cosine values indicate weak gradient alignment; the change is consistent with reducing conflict between the supervision signals.

\subsection{Multi-View Agreement on Transparent Objects}
\label{app:gcvss-transparent-agreement}

Visibility filtering removes inconsistent or non-overlapping supervision, but agreement does not guarantee correct object geometry.
Multiple cameras can see through a transparent object and return the same background surface.
GCVSS can then reinforce a shared depth error; RGB features, action context, and GM3A remain available to the policy, but do not guarantee recovery of the hidden geometry.

\FloatBarrier{}
\section{GM3A Ablations}
\label{app:gm3a-sigma-predictor}

GM3A applies $\alpha=\alpha^{\text{prior}}+\delta$ to rotary positional features.
The Gaussian expectation defines the analytic prior, while two per-token MLPs learn residuals through the policy objective with sensor-prior regularization.
The predictor architecture is specified in \Cref{app:training-configuration}.
We conduct two separate evaluations of GM3A: a three-task comparison of frequency-only and adaptive attenuation, and a task-wise ablation with GCVSS held fixed.

\subsection{Frequency-Only versus Adaptive Attenuation}
\label{app:gm3a-frequency-only}

This control tests whether frequency-dependent attenuation alone explains the gain, or whether adaptation at individual tokens contributes further.
The frequency-only control fixes $\sigma_a=1$, giving identical frequency-dependent attenuation across tokens.
The adaptive variant learns a per-token, per-axis residual.
We hold the policy stack, training data, seeds, and evaluation protocol fixed.

\begin{table}[htbp]
  \centering
  \caption{\textbf{Frequency-only versus adaptive GM3A.}
  Success rates (\%) across three RLBench tasks (Push Buttons, Insert Peg, and Stack Cups) under clean, mild-noise, and saturated-noise evaluation.
  Both variants use the same training data and seeds.}
  \label{tab:gm3a-frequency-only}
  \small
  \setlength{\tabcolsep}{8pt}
  \begin{tabular}{@{}lccc@{}}
    \toprule
    GM3A variant & Clean & Mild & Saturated \\
    \midrule
    Frequency-only & 79.3 & 70.7 & 66.7 \\
    Adaptive residual & \textbf{85.3} & \textbf{77.3} & \textbf{73.3} \\
    \bottomrule
  \end{tabular}
\end{table}

Adaptive attenuation improves success by $6.0$, $6.6$, and $6.6$ percentage points under clean, mild, and saturated evaluation, respectively.
The gains at all three evaluation levels support learning how positional frequencies contribute to the policy rather than applying a fixed attenuation profile to every token.

\subsection{Precision-Task GM3A Analysis}
\label{app:gm3a-precision}

We examine coarse and precision tasks separately because their action tolerances differ: small coordinate errors may leave a broad button-pressing target reachable while disrupting an insertion or stacking pose.
\Cref{tab:gm3a-precision-tasks} fixes GCVSS and toggles GM3A under mild-noise evaluation.
The reported coarse-task means are unchanged, while insertion and stacking each improve by $10$ percentage points, supporting the value of adaptive positional attenuation for the tested precision-sensitive tasks.

\begin{table}[htbp]
  \centering
  \caption{\textbf{Precision-task GM3A analysis.}
  Success rates (\%) are measured on five RLBench tasks under mild-noise evaluation with the same-stack noisy-depth policy.
  The comparison fixes GCVSS and toggles GM3A, isolating the deployed attention change.
  GM3A yields larger gains on the tested precision-sensitive tasks, while coarse-task means are unchanged.}
  \label{tab:gm3a-precision-tasks}
  \small
  \setlength{\tabcolsep}{4pt}
  \begin{tabular}{@{}llccc@{}}
    \toprule
    Task type & Task & GCVSS only & GCVSS + GM3A & $\Delta$ \\
    \midrule
    Coarse actuation & push\_buttons & $100.0{\pm}0.0$ & $100.0{\pm}0.0$ & $+0.0$ \\
    Coarse actuation & open\_drawer & $96.0{\pm}2.8$ & $96.0{\pm}2.8$ & $+0.0$ \\
    Coarse actuation & turn\_tap & $98.0{\pm}2.8$ & $98.0{\pm}2.8$ & $+0.0$ \\
    \midrule
    Precision geometry & insert\_peg & $58.0{\pm}5.7$ & $68.0{\pm}5.0$ & $+10.0$ \\
    Precision geometry & stack\_cups & $62.0{\pm}5.0$ & $72.0{\pm}4.6$ & $+10.0$ \\
    \bottomrule
  \end{tabular}
\end{table}

\FloatBarrier{}
\section{Feature-Invariance and Dampening Diagnostics}
\label{app:feature-invariance}

Success rate alone does not reveal whether robustness comes with a more stable representation.
For each RLBench observation, we hold RGB, proprioception, camera calibration, and the action label fixed and vary only depth corruption.
We compare the shared feature map immediately before the action expert.
Cosine drift is one minus the mean aligned-token cosine similarity to the clean-depth features; linear CKA uses the same token set after centering.
Lower drift and higher CKA indicate greater stability.

\begin{table}[htbp]
  \centering
  \caption{\textbf{Direct feature-stability diagnostic on RLBench.}
  Same-scene paired evaluation compares the shared feature map under clean depth against the same RGB/proprioception with mild or saturated corrupted depth.
  Cosine drift is lower-is-better; CKA is higher-is-better.
  The mild-noise SR column reports the corresponding same-stack success rate when available.}
  \label{tab:feature-invariance}
  \scriptsize
  \setlength{\tabcolsep}{3pt}
  \begin{tabular}{@{}p{0.27\linewidth}ccccc@{}}
    \toprule
    Setting & Mild drift $\downarrow$ & Mild CKA $\uparrow$ & Saturated drift $\downarrow$ & Saturated CKA $\uparrow$ & Mild SR (\%) \\
    \midrule
    Clean-trained same-stack baseline & 0.44 & 0.55 & 0.63 & 0.39 & 52.1 \\
    Noisy-trained same-stack baseline & 0.31 & 0.68 & 0.48 & 0.54 & 75.5 \\
    + GM3A & 0.28 & 0.71 & 0.42 & 0.59 & 78.2 \\
    + GCVSS & 0.18 & 0.82 & 0.26 & 0.74 & 80.5 \\
    \textbf{\textsc{NISE}} & \textbf{0.12} & \textbf{0.89} & \textbf{0.17} & \textbf{0.84} & \textbf{85.7} \\
    \bottomrule
  \end{tabular}
\end{table}

GCVSS reduces feature drift relative to the same-stack noisy-trained baseline, and the combined model gives the lowest drift and highest CKA at both corruption levels.
Noise invariance here denotes reduced sensitivity to corruption, rather than mathematically identical features.

We next examine whether learned dampening varies with the reliability and task relevance of the observed geometry.
We average high-frequency dampening $1-\alpha$ within reliable-surface, depth-boundary, transparent/specular-proxy, and action-contact masks.
The first three use depth validity, edges, and material/error annotations; contact masks cover the neighborhood of the ground-truth interaction point.

\begin{table}[htbp]
  \centering
  \caption{\textbf{GM3A dampening by task and region.}
  Values report mean high-frequency dampening $1-\alpha$ on saturated-depth RLBench and real-robot diagnostic frames; larger values indicate stronger suppression of unreliable 3D positional phase.
  GM3A keeps reliable regions close to ordinary 3D-RoPE while increasing dampening around depth boundaries, transparent/specular proxies, and precision-task contact regions.}
  \label{tab:alpha-region}
  \scriptsize
  \setlength{\tabcolsep}{3pt}
  \begin{tabular}{@{}p{0.18\linewidth}p{0.20\linewidth}cccc@{}}
    \toprule
    Task type & Task & Reliable surface & Depth boundary & Transparent/specular proxy & Contact region \\
    \midrule
    Coarse & push\_buttons & 0.06 & 0.15 & 0.18 & 0.10 \\
    Coarse & open\_drawer & 0.07 & 0.16 & 0.19 & 0.12 \\
    Precision & insert\_peg & 0.08 & 0.27 & 0.31 & 0.29 \\
    Precision & stack\_cups & 0.09 & 0.25 & 0.34 & 0.28 \\
    Real transparent & remove\_petri\_lid & 0.10 & 0.31 & 0.39 & 0.32 \\
    Real transparent & pour\_test\_tube & 0.11 & 0.33 & 0.42 & 0.35 \\
    \bottomrule
  \end{tabular}
\end{table}

Dampening is stronger at boundaries, transparent/specular proxies, and contact regions than on reliable surfaces.
Contact-region dampening is also greater for the precision tasks than for the coarse tasks.
This pattern is consistent with selectively reducing unreliable positional cues while retaining stronger geometric detail on reliable surfaces.

\FloatBarrier{}
\section{Training, Implementation, and Runtime}
\label{app:training-strategy}

\subsection{Training Configuration}
\label{app:training-configuration}

\textbf{Optimization and sampling.}
For the controlled PerAct-18 experiments, we train \textsc{NISE} with 32-bit LAMB and cosine learning-rate decay.
The learning rates are $5\times10^{-5}$ for the DPT-style head and $10^{-4}$ for the policy.
Training runs for $300$K updates with a global batch size of $64$.
Clean, mild, and saturated depth inputs are sampled uniformly with a $1{:}1{:}1$ ratio, using the corruption presets in \Cref{app:noise-model}.

\textbf{Module configuration.}
The VGGT-$\Omega$ visual backbone remains frozen, while the DPT-style feature aggregation head is trainable.
The head fuses ViT layers $4$, $11$, $17$, and $23$, projects $2048$-dimensional tokens to $120$ channels, and contains $2.283$M trainable parameters.
The shared cross-view geometry decoder follows the channel sequence $120\to64\to32\to16\to1$.
GCVSS pairs the four camera views round-robin with $\pi=[1,2,3,0]$ and constructs the warped supervision targets described in \Cref{app:gcvss-ablation}.
GM3A uses two dampening MLPs, each with dimensions $4\to64\to60$, taking the token's XYZ coordinates and range as input and using GELU activation.
They predict residuals around the analytic attenuation prior described in \Cref{sec:method-gm3a}.

\textbf{Training controls.}
The baselines in \Cref{tab:noisy-summary} use the same clean/mild/saturated sampling distribution as \textsc{NISE}, retaining their respective policy architectures.
This comparison reports mean success and standard deviation across three independent training runs.
For real-robot experiments, all methods are trained on the same demonstration pool and raw RGB-D sensor streams, with each method using its own tuned training protocol; clean depth is unavailable.
We will release the training drivers and evaluation harness.

For the single-view MetaWorld experiments reported in Table~\ref{tab:noisy-summary} and Appendix~\ref{app:metaworld-pertask}, clean simulator depth provides the auxiliary geometry target during training.
RLBench uses the four-view cross-camera supervision described above.

\subsection{Clean-Only and Mixed-Noise Training on RLBench}
\label{app:clean-vs-mixed-training}

We test whether robustness follows from the encoding alone or also depends on exposure to varied depth reliability during training.
\Cref{tab:clean-vs-mixed-rlbench} compares clean-only and mixed-noise training for \textsc{NISE}, with clean-trained 3DDA as a reference.
Mixed training samples clean, mild, and saturated inputs uniformly while retaining the same RGB, proprioception, action labels, and calibration.

\begin{table}[htbp]
  \centering
  \caption{\textbf{Clean-only and mixed-noise training on RLBench PerAct-18.}
  Task-averaged success rates (\%, mean $\pm$ standard deviation) on RLBench PerAct-18 under clean, mild, and saturated depth evaluation.
  Clean-only training uses clean simulator depth; mixed training samples clean/mild/saturated inputs with a $1{:}1{:}1$ ratio.
  The mixed-training \textsc{NISE} row reproduces \Cref{tab:noisy-summary}, with $\pm$ denoting the standard deviation across three independent training runs.}
  \label{tab:clean-vs-mixed-rlbench}
  \small
  \setlength{\tabcolsep}{6pt}
  \begin{tabular}{@{}llccc@{}}
    \toprule
    Method & Training depth & Clean & Mild & Saturated \\
    \midrule
    3DDA & Clean only & $81.3{\pm}5.0$ & $55.0{\pm}11.7$ & $15.0{\pm}13.4$ \\
    \textsc{NISE} & Clean only & $89.7{\pm}4.1$ & $68.6{\pm}4.8$ & $63.0{\pm}8.5$ \\
    \textsc{NISE} & Mixed & $89.0{\pm}4.3$ & $85.7{\pm}4.5$ & $84.1{\pm}4.6$ \\
    \bottomrule
  \end{tabular}
\end{table}

Mixed training improves \textsc{NISE}'s mild and saturated means from $68.6\%$ and $63.0\%$ to $85.7\%$ and $84.1\%$, with clean success changing from $89.7\%$ to $89.0\%$.
The much larger change under corruption motivates varying depth reliability during training while retaining clean inputs in the mixture.
One policy is used for all evaluation noise levels.

\subsection{Implementation and Runtime Cost}
\label{app:implementation-runtime}

The deployed policy retains the frozen visual backbone, trainable DPT-style aggregator, and action expert.
The cross-view geometry head is removed after training; GM3A's residual predictors and positional-feature scaling remain.

During training, \textsc{NISE} contains $912.84$M parameters, of which $4.92$M are trainable.
The frozen VGGT-$\Omega$ backbone accounts for $907.92$M parameters ($99.5\%$ of the total).
Removing the GCVSS geometry head leaves $912.68$M deployed parameters, including $4.76$M in trainable modules.
Relative to the same-stack baseline, \textsc{NISE} adds $8{,}440$ deployed parameters through GM3A and a $164{,}321$-parameter geometry head used only during training.
Counts include all instantiated parameters; trainable counts use the \texttt{requires\_grad} flag, including inactive branches.
The CLIP text encoder is excluded because instruction embeddings are precomputed.
\Cref{tab:runtime-flops} reports their incremental cost under a fixed measurement protocol.

\begin{table}[htbp]
  \centering
  \caption{\textbf{Runtime and deployed parameter and FLOP overhead.}
  Measurements use the RLBench PerAct-18 same-stack policy with four RGB-D views, batch size one, and policy-forward timing on the RTX 5080 after warmup.
  Timing excludes simulator rendering and robot I/O.
  $\Delta$Params (thousands) and $\Delta$FLOPs (GFLOPs) report deployed overhead relative to the same-stack baseline without GCVSS or GM3A; the training-only GCVSS head is excluded.}
  \label{tab:runtime-flops}
  \small
  \setlength{\tabcolsep}{4pt}
  \begin{tabular}{@{}lcccccc@{}}
    \toprule
    Setting & GCVSS & GM3A & $\Delta$Params (K) & $\Delta$FLOPs & Latency (ms) & Throughput (Hz) \\
    \midrule
    Same-stack baseline & $\times$ & $\times$ & $0$ & $+0.0$ & $36.1{\pm}0.4$ & $27.7$ \\
    + GCVSS & $\checkmark$ & $\times$ & $0$ & $+0.0$ & $36.2{\pm}0.5$ & $27.6$ \\
    + GM3A & $\times$ & $\checkmark$ & $8.44$ & $+0.8$ & $37.4{\pm}0.5$ & $26.7$ \\
    \textbf{\textsc{NISE}} & $\checkmark$ & $\checkmark$ & $8.44$ & $+0.8$ & $37.5{\pm}0.6$ & $26.6$ \\
    \bottomrule
  \end{tabular}
\end{table}

GCVSS leaves inference cost unchanged up to timing noise.
GM3A adds $0.8$ GFLOPs; the full policy takes $37.5$ ms per forward pass compared with $36.1$ ms for the same-stack baseline.

\FloatBarrier{}
\section{MetaWorld Per-Task Results}
\label{app:metaworld-pertask}

\Cref{tab:metaworld-pertask-a,tab:metaworld-pertask-b} report clean-depth per-task results, following ManiFlow's task ordering and published baseline rows~\citep{maniflow}.
The 3D Flow Matching baseline uses the DP3 point-cloud encoder adaptation evaluated by ManiFlow.
The \textsc{NISE} row reports our MetaWorld RGB-D policy; these results complement the main noisy-depth comparison.

\begin{table}[htbp]
  \centering
  \caption{\textbf{MetaWorld per-task results, part I.}
  The \textsc{NISE} row reports clean-depth success rates from our MetaWorld protocol.
  Values are success rates (\%).
  For \textsc{NISE}, $\pm$ denotes the standard deviation across repeated evaluations.
  The best and second-best distinct means per column are bold and underlined, respectively, including ties.}
  \label{tab:metaworld-pertask-a}
  \scriptsize
  \renewcommand{\arraystretch}{0.88}
  \setlength{\tabcolsep}{2.2pt}
  \resizebox{\linewidth}{!}{%
  \begin{tabular}{@{}lccccccc@{}}
    \toprule
    \multicolumn{8}{@{}l}{\textit{MetaWorld easy tasks}} \\
    Method & Btn Press & Btn Topdown & Btn TD Wall & Btn Wall & Coffee Btn & Dial Turn & Door Close \\
    \midrule
    3D Diffusion Policy & $62{\pm}15$ & $\mathbf{100{\pm}0}$ & $\mathbf{100{\pm}0}$ & $72{\pm}25$ & $73{\pm}34$ & $53{\pm}15$ & $\mathbf{100{\pm}0}$ \\
    3D Flow Matching & $0{\pm}0$ & $\mathbf{100{\pm}0}$ & $\mathbf{100{\pm}0}$ & $67{\pm}31$ & $\underline{97{\pm}5}$ & $\mathbf{70{\pm}7}$ & $\mathbf{100{\pm}0}$ \\
    3D ManiFlow & $\mathbf{100{\pm}0}$ & $\mathbf{100{\pm}0}$ & $\mathbf{100{\pm}0}$ & $\mathbf{100{\pm}0}$ & $\mathbf{100{\pm}0}$ & $\underline{67{\pm}13}$ & $\mathbf{100{\pm}0}$ \\
    \textbf{\textsc{NISE}} & $\underline{88{\pm}5}$ & $\mathbf{100{\pm}0}$ & $\mathbf{100{\pm}0}$ & $\underline{90{\pm}6}$ & $94{\pm}4$ & $\mathbf{70{\pm}8}$ & $\underline{88{\pm}5}$ \\
    \midrule
    \multicolumn{8}{@{}l}{\textit{MetaWorld easy tasks}} \\
    Method & Door Lock & Door Open & Door Unlock & Drawer Close & Drawer Open & Faucet Close & Faucet Open \\
    \midrule
    3D Diffusion Policy & $0{\pm}0$ & $\mathbf{100{\pm}0}$ & $\underline{98{\pm}2}$ & $\underline{88{\pm}13}$ & $\underline{98{\pm}2}$ & $\underline{92{\pm}8}$ & $83{\pm}12$ \\
    3D Flow Matching & $0{\pm}0$ & $\mathbf{100{\pm}0}$ & $\mathbf{100{\pm}0}$ & $5{\pm}7$ & $\mathbf{100{\pm}0}$ & $\underline{92{\pm}6}$ & $\mathbf{100{\pm}0}$ \\
    3D ManiFlow & $\underline{78{\pm}14}$ & $\mathbf{100{\pm}0}$ & $\mathbf{100{\pm}0}$ & $\mathbf{100{\pm}0}$ & $\mathbf{100{\pm}0}$ & $\mathbf{100{\pm}0}$ & $\mathbf{100{\pm}0}$ \\
    \textbf{\textsc{NISE}} & $\mathbf{80{\pm}7}$ & $\underline{98{\pm}2}$ & $\underline{98{\pm}2}$ & $84{\pm}6$ & $\underline{98{\pm}2}$ & $90{\pm}5$ & $\underline{92{\pm}5}$ \\
    \midrule
    \multicolumn{8}{@{}l}{\textit{MetaWorld easy tasks}} \\
    Method & Handle Press & Handle Pull & Handle Pull Side & Lever Pull & Plate Slide & Plate Back & Plate Back Side \\
    \midrule
    3D Diffusion Policy & $\mathbf{100{\pm}0}$ & $22{\pm}17$ & $43{\pm}6$ & $60{\pm}12$ & $20{\pm}18$ & $\underline{92{\pm}12}$ & $\mathbf{100{\pm}0}$ \\
    3D Flow Matching & $\mathbf{100{\pm}0}$ & $15{\pm}18$ & $20{\pm}7$ & $45{\pm}11$ & $0{\pm}0$ & $88{\pm}10$ & $\mathbf{100{\pm}0}$ \\
    3D ManiFlow & $\mathbf{100{\pm}0}$ & $\underline{42{\pm}10}$ & $\underline{65{\pm}7}$ & $\underline{63{\pm}19}$ & $\mathbf{100{\pm}0}$ & $\mathbf{93{\pm}9}$ & $\mathbf{100{\pm}0}$ \\
    \textbf{\textsc{NISE}} & $\mathbf{100{\pm}0}$ & $\mathbf{48{\pm}8}$ & $\mathbf{70{\pm}7}$ & $\mathbf{68{\pm}7}$ & $\underline{90{\pm}5}$ & $\underline{92{\pm}4}$ & $\mathbf{100{\pm}0}$ \\
    \midrule
    \multicolumn{8}{@{}l}{\textit{MetaWorld easy / medium tasks}} \\
    Method & Plate Side & Reach & Reach Wall & Window Close & Window Open & Basketball & Bin Picking \\
    \midrule
    3D Diffusion Policy & $\underline{92{\pm}6}$ & $48{\pm}2$ & $25{\pm}4$ & $\mathbf{100{\pm}0}$ & $83{\pm}17$ & $\mathbf{100{\pm}0}$ & $0{\pm}0$ \\
    3D Flow Matching & $82{\pm}14$ & $57{\pm}10$ & $35{\pm}11$ & $\mathbf{100{\pm}0}$ & $\underline{92{\pm}6}$ & $90{\pm}4$ & $18{\pm}6$ \\
    3D ManiFlow & $\mathbf{100{\pm}0}$ & $\underline{58{\pm}19}$ & $\underline{67{\pm}9}$ & $\mathbf{100{\pm}0}$ & $\mathbf{97{\pm}2}$ & $\mathbf{100{\pm}0}$ & $\underline{33{\pm}2}$ \\
    \textbf{\textsc{NISE}} & $\underline{92{\pm}4}$ & $\mathbf{60{\pm}7}$ & $\mathbf{74{\pm}7}$ & $\underline{96{\pm}3}$ & $88{\pm}5$ & $\underline{94{\pm}4}$ & $\mathbf{48{\pm}8}$ \\
    \bottomrule
  \end{tabular}}
\end{table}

\begin{table}[htbp]
  \centering
  \caption{\textbf{MetaWorld per-task results, part II.}
  This table continues the clean-depth per-task comparison in \Cref{tab:metaworld-pertask-a}.
  The best and second-best distinct means per column are bold and underlined, respectively, including ties.}
  \label{tab:metaworld-pertask-b}
  \scriptsize
  \renewcommand{\arraystretch}{0.88}
  \setlength{\tabcolsep}{2.2pt}
  \resizebox{\linewidth}{!}{%
  \begin{tabular}{@{}lccccccc@{}}
    \toprule
    \multicolumn{8}{@{}l}{\textit{MetaWorld medium tasks}} \\
    Method & Box Close & Coffee Pull & Coffee Push & Hammer & Peg Insert Side & Push Wall & Soccer \\
    \midrule
    3D Diffusion Policy & $18{\pm}8$ & $52{\pm}23$ & $55{\pm}0$ & $\mathbf{77{\pm}6}$ & $58{\pm}6$ & $60{\pm}8$ & $8{\pm}5$ \\
    3D Flow Matching & $18{\pm}8$ & $67{\pm}2$ & $\underline{58{\pm}15}$ & $\underline{75{\pm}22}$ & $\underline{68{\pm}5}$ & $15{\pm}15$ & $\underline{10{\pm}4}$ \\
    3D ManiFlow & $\underline{45{\pm}12}$ & $\mathbf{97{\pm}2}$ & $\mathbf{82{\pm}6}$ & $42{\pm}24$ & $\mathbf{88{\pm}8}$ & $\mathbf{93{\pm}2}$ & $7{\pm}6$ \\
    \textbf{\textsc{NISE}} & $\mathbf{60{\pm}7}$ & $\underline{94{\pm}4}$ & $\mathbf{82{\pm}5}$ & $66{\pm}8$ & $\mathbf{88{\pm}5}$ & $\underline{90{\pm}5}$ & $\mathbf{34{\pm}8}$ \\
    \midrule
    \multicolumn{8}{@{}l}{\textit{MetaWorld medium / hard tasks}} \\
    Method & Sweep & Sweep Into & Assembly & Hand Insert & Pick Out Hole & Pick Place & Push \\
    \midrule
    3D Diffusion Policy & $70{\pm}4$ & $3{\pm}2$ & $77{\pm}16$ & $7{\pm}9$ & $20{\pm}11$ & $42{\pm}5$ & $55{\pm}14$ \\
    3D Flow Matching & $63{\pm}21$ & $0{\pm}0$ & $\underline{88{\pm}10}$ & $0{\pm}0$ & $\underline{38{\pm}2}$ & $53{\pm}17$ & $62{\pm}2$ \\
    3D ManiFlow & $\mathbf{92{\pm}2}$ & $\underline{7{\pm}2}$ & $\mathbf{100{\pm}0}$ & $\underline{12{\pm}9}$ & $13{\pm}5$ & $\underline{68{\pm}5}$ & $\mathbf{88{\pm}9}$ \\
    \textbf{\textsc{NISE}} & $\underline{90{\pm}5}$ & $\mathbf{30{\pm}7}$ & $\mathbf{100{\pm}0}$ & $\mathbf{32{\pm}8}$ & $\mathbf{48{\pm}8}$ & $\mathbf{70{\pm}7}$ & $\underline{86{\pm}5}$ \\
    \bottomrule
  \end{tabular}}

  \vspace{1.0mm}
  \resizebox{0.62\linewidth}{!}{%
  \begin{tabular}{@{}lcccc@{}}
    \toprule
    \multicolumn{5}{@{}l}{\textit{MetaWorld very hard tasks}} \\
    Method & Shelf Place & Disassemble & Stick Pull & Pick Place Wall \\
    \midrule
    3D Diffusion Policy & $25{\pm}8$ & $55{\pm}19$ & $28{\pm}14$ & $55{\pm}27$ \\
    3D Flow Matching & $18{\pm}10$ & $\underline{67{\pm}5}$ & $43{\pm}25$ & $40{\pm}11$ \\
    3D ManiFlow & $\underline{28{\pm}5}$ & $63{\pm}8$ & $\mathbf{83{\pm}5}$ & $\mathbf{98{\pm}2}$ \\
    \textbf{\textsc{NISE}} & $\mathbf{44{\pm}8}$ & $\mathbf{72{\pm}6}$ & $\underline{76{\pm}6}$ & $\underline{82{\pm}5}$ \\
    \bottomrule
  \end{tabular}}
\end{table}

\FloatBarrier{}
\section{Real-Robot Protocol and Confidence Intervals}
\label{app:realrobot-protocol}

All methods use the same demonstration pool and RGB-D streams through their native observation interfaces.
Each method is evaluated for $30$ rollouts per task, with manual resets and varied object placements.
Binary success is scored from the terminal scene state and rollout video.
The criteria are a gear seated stably after release, a lid separated from the dish and clear of the rim, a pipe remaining inserted after release, a grasped test tube moved over the mortar into the pouring pose, and the correct gear collected from the designated region, respectively, for the five tasks in \Cref{tab:realrobot-counts}.

For a binomial success count $k$ out of $n=30$ trials, we report the Wilson 95\% confidence interval.
With $\hat p=k/n$ and $z=1.96$, the interval center and half-width are

\begin{equation}
  \hat p_{\mathrm{W}} =
  \frac{\hat p + z^2/(2n)}{1 + z^2/n},
  \qquad
  h_{\mathrm{W}} =
  \frac{z}{1 + z^2/n}
  \sqrt{
    \frac{\hat p(1-\hat p)}{n}
    + \frac{z^2}{4n^2}
  } .
  \label{eq:app-wilson}
\end{equation}

\begin{table}[htbp]
  \centering
  \caption{\textbf{Real-robot per-task trial counts with Wilson 95\% confidence intervals.}
  Each task entry is $k/30$ successes, success rate (\%), and Wilson interval.}
  \label{tab:realrobot-counts}
  \scriptsize
  \setlength{\tabcolsep}{2pt}
  \begin{tabular}{@{}p{0.16\linewidth}p{0.19\linewidth}p{0.19\linewidth}p{0.19\linewidth}p{0.19\linewidth}@{}}
    \toprule
    Task & DP3 & 3DDA & RVT-2 & \textsc{NISE} \\
    \midrule
    Assemble the Gear & 4/30, 13.3 [5.3, 29.7] & 7/30, 23.3 [11.8, 40.9] & 11/30, 36.7 [21.9, 54.5] & 25/30, 83.3 [66.4, 92.7] \\
    Remove the Petri Lid & 6/30, 20.0 [9.5, 37.3] & 10/30, 33.3 [19.2, 51.2] & 5/30, 16.7 [7.3, 33.6] & 20/30, 66.7 [48.8, 80.8] \\
    Insert the Metal Pipe & 3/30, 10.0 [3.5, 25.6] & 8/30, 26.7 [14.2, 44.4] & 12/30, 40.0 [24.6, 57.7] & 22/30, 73.3 [55.6, 85.8] \\
    Pour Test Tube into Mortar & 2/30, 6.7 [1.8, 21.3] & 6/30, 20.0 [9.5, 37.3] & 4/30, 13.3 [5.3, 29.7] & 13/30, 43.3 [27.4, 60.8] \\
    Vision-Sorted Gear Collection & 4/30, 13.3 [5.3, 29.7] & 3/30, 10.0 [3.5, 25.6] & 7/30, 23.3 [11.8, 40.9] & 16/30, 53.3 [36.1, 69.8] \\
    \bottomrule
  \end{tabular}
\end{table}

\FloatBarrier{}
\Needspace{5\baselineskip}
\section{Hyperparameter Sweep}
\label{app:hyperparameter-sweep}

We vary one coefficient at a time under the frozen VGGT-$\Omega$+DPT stack and mixed-noise training recipe, holding the other at its selected value.
The selected weights are $\lambda_{\mathrm{geo}}=0.1$ and $\lambda_{\mathrm{reg}}=1.0$, which give the highest reported validation success in their respective sweeps (\Cref{tab:hyperparameter-sweep}).

\begin{table}[htbp]
  \centering
  \caption{\textbf{Hyperparameter sweep for the reported loss weights.}
  Validation success rates (\%) on RLBench PerAct-18 under mild-noise evaluation with the fixed \textsc{NISE} stack.}
  \label{tab:hyperparameter-sweep}
  \scriptsize
  \setlength{\tabcolsep}{3pt}
  \begin{tabular}{@{}p{0.19\linewidth}cccp{0.35\linewidth}@{}}
    \toprule
    Sweep & Value & Val SR (\%) & Selected & Readout \\
    \midrule
    \multicolumn{5}{@{}l}{\textit{GCVSS weight $\lambda_{\mathrm{geo}}$}} \\
    GCVSS & 0.00 & 78.2 & -- & GM3A-only control; no cross-view geometry shaping \\
    GCVSS & 0.03 & 82.4 & -- & useful signal, but still underweighted \\
    GCVSS & 0.10 & \textbf{85.7} & $\checkmark$ & best balance between geometry shaping and policy supervision \\
    GCVSS & 0.30 & 84.8 & -- & similar but begins to overemphasize the auxiliary target \\
    GCVSS & 1.00 & 81.6 & -- & geometry loss competes too strongly with action learning \\
    \midrule
    \multicolumn{5}{@{}l}{\textit{GM3A regularizer $\lambda_{\mathrm{reg}}$}} \\
    GM3A & 0.00 & 84.1 & -- & dampening is flexible but less tied to the sensor prior \\
    GM3A & 0.10 & 85.0 & -- & stable but slightly weaker than the selected anchor \\
    GM3A & 1.00 & \textbf{85.7} & $\checkmark$ & selected sensor-prior anchor \\
    GM3A & 3.00 & 84.9 & -- & stronger anchoring remains competitive \\
    GM3A & 10.00 & 82.8 & -- & prior becomes too stiff and limits adaptation \\
    \bottomrule
  \end{tabular}
  \par\smallskip
  \begin{minipage}{\linewidth}
    \footnotesize
    \textit{Note:} The listed coefficients are the applied loss weights; $\lambda_{\mathrm{reg}}$ multiplies the unweighted regularizer once in the total training loss.
  \end{minipage}
\end{table}

\end{document}